\documentclass[10pt,twocolumn,letterpaper]{article}

\usepackage[pagenumbers]{cvpr} 

\usepackage{graphicx}
\usepackage{amsmath}
\usepackage{amssymb}
\usepackage{booktabs}

\usepackage{multirow}
\usepackage{booktabs}
\usepackage{amsmath,amssymb}
\usepackage{wrapfig}
\usepackage{color}

\usepackage[pagebackref,breaklinks,colorlinks]{hyperref}

\hypersetup{
    colorlinks=true,
    linkcolor=red,
    filecolor=magenta,      
    urlcolor=magenta,
}

\usepackage[capitalize]{cleveref}
\crefname{section}{Sec.}{Secs.}
\Crefname{section}{Section}{Sections}
\Crefname{table}{Table}{Tables}
\crefname{table}{Tab.}{Tabs.}

\def\cvprPaperID{1831} 
\def\confName{CVPR}
\def\confYear{2022}

\newcommand{\parsection}[1]{\noindent\textbf{#1}}

\begin{document}


\title{Mask Transfiner for High-Quality Instance Segmentation}

\author{
 Lei Ke$^{1,2}$\hspace{0.35cm}Martin Danelljan$^1$\hspace{0.35cm}Xia Li$^1$\hspace{0.35cm}Yu-Wing Tai$^3$\hspace{0.35cm}Chi-Keung Tang$^2$\hspace{0.35cm}Fisher Yu$^1$\\
 $^1$ETH Z{\"u}rich\hspace{1.5cm}$^2$HKUST\hspace{1.5cm}$^3$Kuaishou Technology \\
 }
  
\maketitle

\begin{abstract}
   Two-stage and query-based instance segmentation methods have achieved remarkable results. However, their segmented masks are still very coarse.
   In this paper, we present {\em Mask Transfiner} for high-quality and efficient instance segmentation. Instead of operating on regular dense tensors, our Mask Transfiner decomposes and represents the image regions as a quadtree. Our transformer-based approach only processes detected error-prone tree nodes and self-corrects their errors in parallel. While these sparse pixels only constitute a small proportion of the total number, they are critical to the final mask quality. This allows Mask Transfiner to predict highly accurate instance masks, at a low computational cost.
   Extensive experiments demonstrate that Mask Transfiner outperforms current instance segmentation methods on three popular benchmarks, significantly improving both two-stage and query-based frameworks by a large margin of +3.0 mask AP on COCO and BDD100K, and +6.6 boundary AP on Cityscapes. Our code and trained models will be available at \url{http://vis.xyz/pub/transfiner}. 
\end{abstract}
\vspace{-0.2in}

\section{Introduction}
\label{sec:intro}

Advancements in image instance segmentation has largely been driven by the developments of powerful object detection paradigms. Approaches based on Mask R-CNN~\cite{he2017mask,liu2018path,huang2019mask,ChengWHL20,kirillov2020pointrend} and more recently DETR~\cite{dong2021solq,QueryInst,hu2021ISTR} have achieved ever increasing performance on, for instance, the COCO challenge~\cite{lin2014microsoft}. While these methods excel in detection and localization of objects, the problem of efficiently predicting highly accurate segmentation masks has so far remained elusive.


As shown in Figure~\ref{tab:gap}, there is still a significant gap between the bounding box and segmentation performance of the recent state-of-the-art methods, especially for the recent query-based methods. This strongly indicates that improvements in mask quality has not kept pace with the advancements detection capability. In Figure~\ref{fig:fig1}, the predicted masks of previous methods are very coarse, most often over-smoothing object boundaries. In fact, efficient and accurate mask prediction is highly challenging, due to the need for high-resolution deep features, which demands large computational and memory costs~\cite{wang2020deep}. 

\begin{figure}[!t]
	\centering
	\vspace{-0.25in}
	\includegraphics[width=1.0\linewidth]{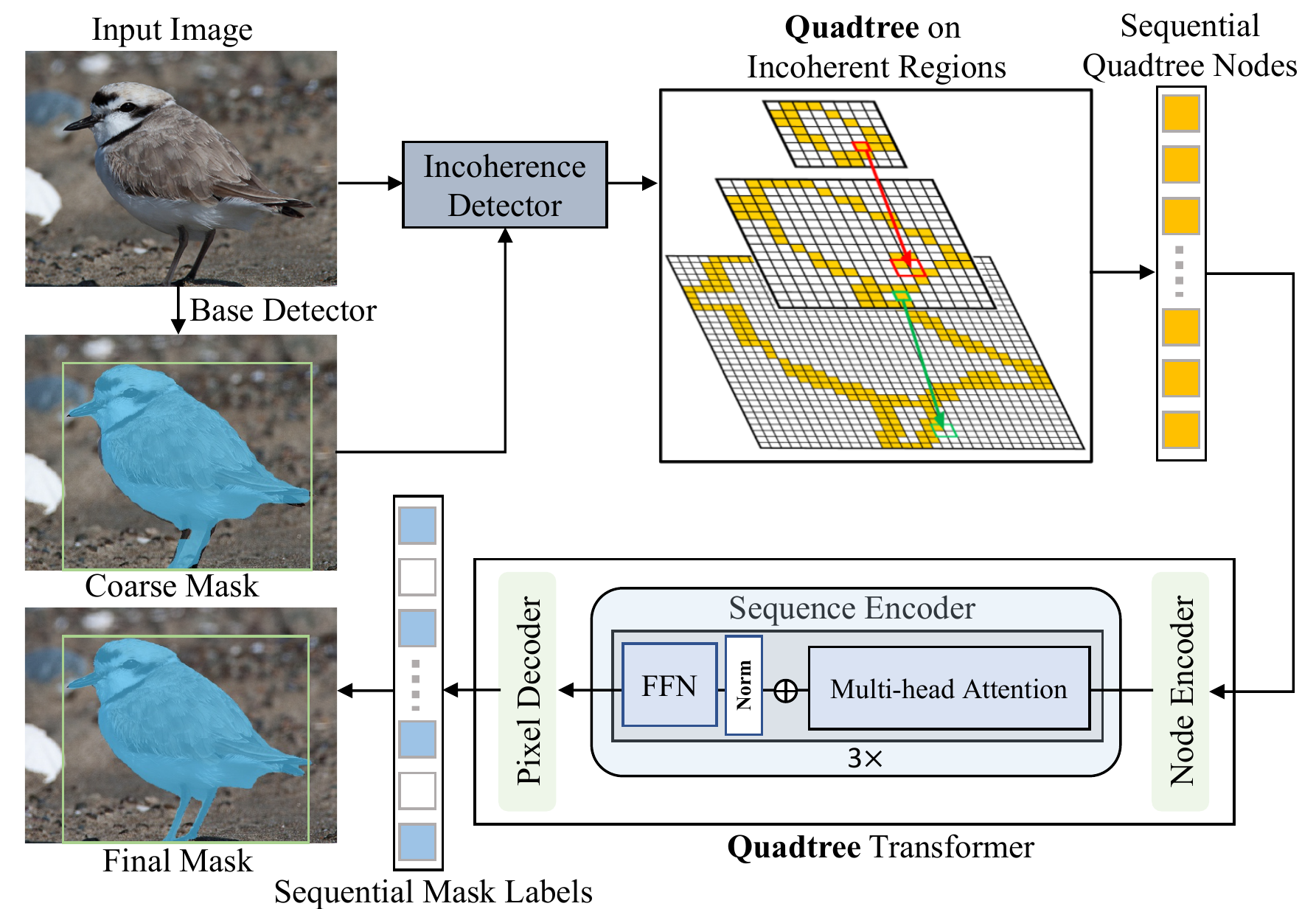}
	\vspace{-0.28in}
	\caption{We propose Mask Transfiner for high-quality instance segmentation. It first builds a quadtree based on the sparse incoherent regions on the RoI pyramid and then jointly refines all tree nodes using the refinement transformer with quadtree attention.}
	\label{fig:new_teaser}
	\vspace{-0.26in}
\end{figure}

To address these issues, we propose Mask Transfiner, an efficient transformer-based approach for high-quality instance segmentation. In Figure~\ref{fig:new_teaser}, our approach first identifies error-prone regions, which are mostly strewn along object boundaries or in high-frequency regions. To this end, our network learns to detect \emph{incoherent regions}, defined by the loss of information when downsampling mask itself. 
These incoherent pixels are sparsely located, consisting only of a small portion of the total pixels. However, as they are shown to be critical to the final segmentation performance, it allows us to only process small parts of the high-resolution feature maps in the refinement process.
Thus, we build a hierarchical quadtree~\cite{finkel1974quad} to represent and process the incoherent image pixels at multiple scales.

\begin{figure*}[!t]
    \vspace{-0.1in}
	\centering
    \includegraphics[width=1.0\linewidth]{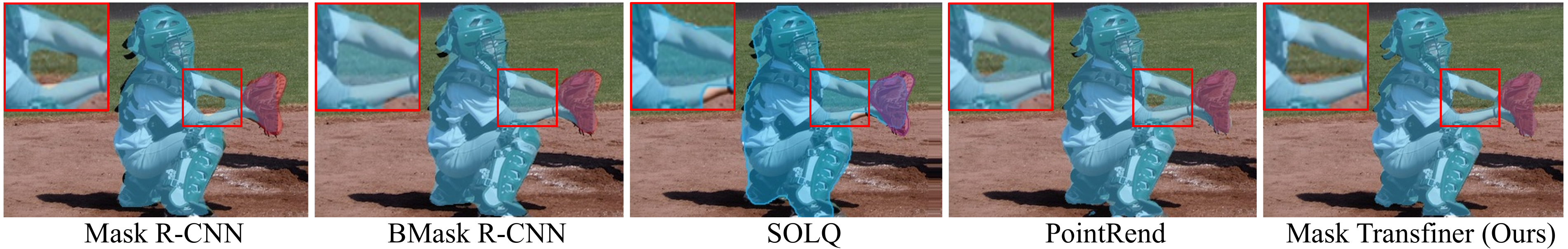}
	\vspace{-0.25in}
	\caption{Instance Segmentation on COCO~\cite{lin2014microsoft} validation set by a) Mask R-CNN~\cite{he2017mask}, b) BMask R-CNN~\cite{ChengWHL20}, c) SOLQ~\cite{dong2021solq}, d) PointRend~\cite{kirillov2020pointrend}, g) Mask Transfiner (Ours) using R50-FPN as backbone, where Mask Transfiner produces significantly more detailed results at high-frequency image regions by replacing Mask R-CNN's default mask head. Zoom in for better view.}
    \label{fig:fig1}
    \vspace{-0.2in}
\end{figure*}

\begin{figure}[!t]
	\centering
	\includegraphics[width=0.95\linewidth]{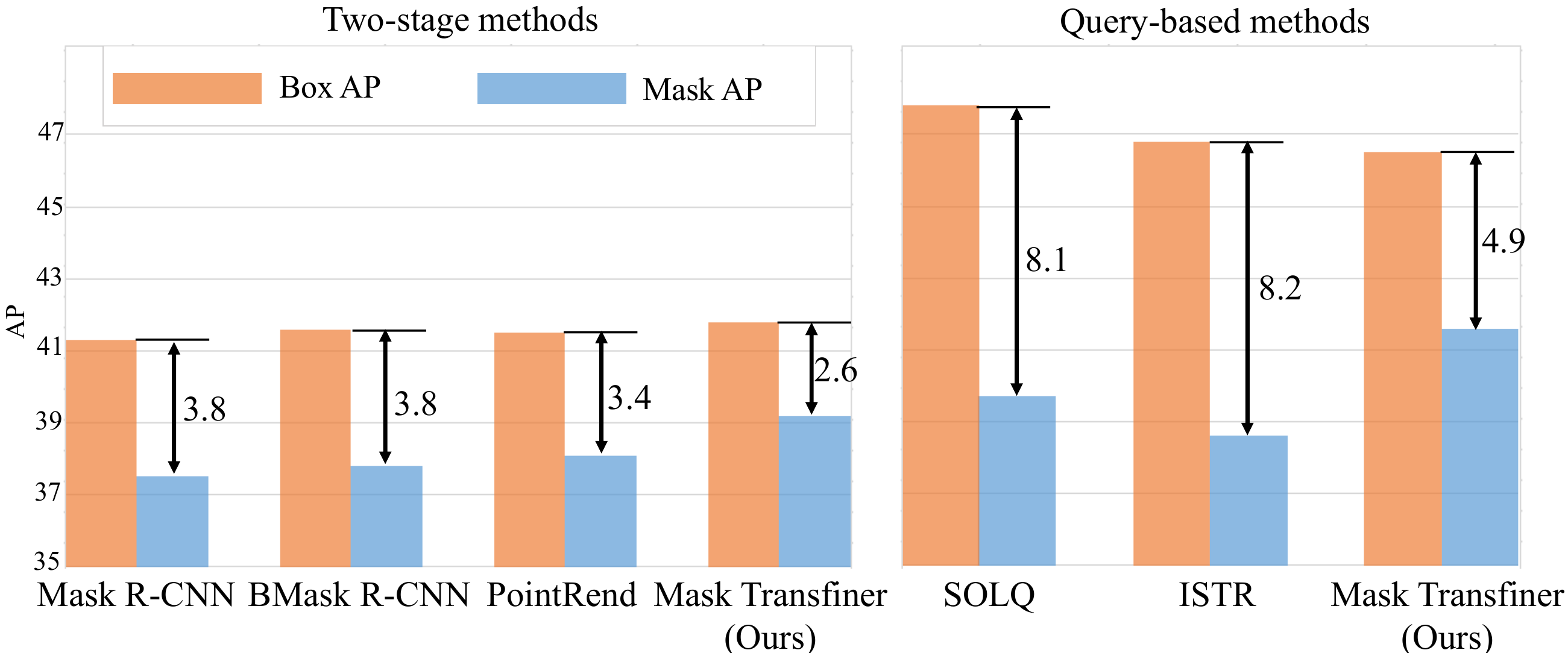}
	\vspace{-0.1in}
	\caption{The performance gap between object detection and segmentation for instance segmentation models on COCO \textit{test-dev} set using R50-FPN as backbone. Detailed comparisons are in Table~\ref{table:fully}.}
	\label{tab:gap}
	\vspace{-0.15in}
\end{figure}



To refine the mask labels of the incoherent quadtree nodes, we design an refinement network based on the transformer instead of standard convolutional networks because they require operating on uniform grids. Our transformer has three modules: node encoder, sequence encoder and pixel decoder. The node encoder first enriches the feature embedding for each incoherent point. The sequence encoder then takes these encoded feature vectors across multiple quadtree levels as input queries. Finally, the pixel decoder predicts their corresponding mask labels. Comparing to MLP~\cite{kirillov2020pointrend}, the sequential representation and multi-head attention enables Mask Transfiner to flexibly takes as input sparse feature points across levels in parallel, models their pixel-wise relations, and then propagates information among them even in a long distance range. 

We extensively analyze our approach on COCO, Cityscapes and BDD100K benchmarks, where quantitative and qualitative results show that Mask Transfiner not only outperforms existing two-stage and query-based methods, but also is efficient in computation and memory cost compared to standard transformer usages. We establish a new state-of-the-art result on COCO \textit{test-dev} of 41.6 AP$^\text{Mask}$ using ResNet-50, outperforming most recent SOLQ~\cite{dong2021solq} and QueryInst~\cite{QueryInst} by a significant margin.
\vspace{-0.1in}

\section{Related Work}
\label{sec:formatting}

\parsection{Instance Segmentation}
Two-stage instance segmentation methods~\cite{li2017fully,he2017mask,cai2018cascade,chen2019hybrid,chen2019tensormask,huang2019mask,pcan,liang2020polytransform} first detects bounding boxes and then performing segmentation in each RoI region.
Mask R-CNN~\cite{he2017mask} extends Faster R-CNN~\cite{ren2015faster} with an FCN branch.
The follow-up works~\cite{chen2018masklab,ChengWHL20,liu2018path,ke2021bcnet,RSLoss} also contribute to the family of Mask R-CNN models.
One-stage methods~\cite{chen2020blendmask, chen2019tensormask, kuo2019shapemask,lee2019centermask} and kernel-based method~\cite{zhang2021k}, such as
PolarMask~\cite{xie2019polarmask}, YOLOACT~\cite{bolya2019yolact}, and SOLO~\cite{wang2019solo,wang2020solov2} remove the proposal generation and feature re-pooling steps, achieving comparable results with higher efficiency.

Query-based instance segmentation methods~\cite{QueryInst,dong2021solq,wang2020end,guo2021sotr,hu2021ISTR},
which are inspired by DETR~\cite{carion2020end}, have emerged very recently by treating segmentation as a set prediction problem.
These methods use queries to represent the interested objects and jointly perform classification, detection and mask regression on them. In~\cite{dong2021solq,hu2021ISTR}, the object masks are compressed as encoding vectors using DCT or PCA algorithms, while QueryInst~\cite{QueryInst} adopts dynamic mask heads with mask information flow. However, the large gaps between the detection and segmentation performance in Figure~\ref{tab:gap} reveals that the mask quality produced by these query-based methods are still unsatisfactory. In contrast to the above methods, Mask Transfiner is targeted for high-quality instance segmentation. In our efficient transformer the input queries are incoherent pixels nodes, instead of representing the objects. Our method is applicable to and effective in both the two-stage and query-based frameworks. 

\parsection{Refinement for Instance Segmentation} Most existing works on instance segmentation refinement rely on specially designed convolutional networks~\cite{tang2021look,refinemask} or MLPs~\cite{kirillov2020pointrend}. PointRend~\cite{kirillov2020pointrend} samples feature points with low-confidence scores and refines their labels with a shared MLP, where the selected points are determined by the coarse predictions of the Mask R-CNN. RefineMask~\cite{refinemask} incorporates fine-grained features with an additional semantic head as the guidance. 
The post-processing method BPR~\cite{tang2021look} crops boundary patches of images and initial masks as input and use~\cite{wang2020deep} for segmentation. Notably some methods~\cite{takikawa2019gated,wang2020deep,yuan2020segfix,cheng2020cascadepsp} focus on refining semantic segmentation details. However, it is much more challenging for instance segmentation due to the more complex segmentation setting, with varying number of objects per image and the requirement of delineating similar and overlapping objects.   
\vspace{-0.15in}

Compared to these refinement methods, Mask Transfiner is an end-to-end instance segmentation method, using a transformer for correcting errors. The regions to be refined are predicted using a lightweight FCN, instead of non-deterministic sampling based on mask scores~\cite{kirillov2020pointrend}. Different from the MLP in~\cite{kirillov2020pointrend}, the sequential and hierarchical input representation enables Mask Transfiner to efficiently take non-local sparse feature points as input queries, where the strong global processing of transformers is a natural fit for our quadtree structure.




\vspace{-0.1in}
\section{Mask Transfiner}

We propose an approach to efficiently tackle high-quality instance segmentation. The overall architecture of Mask Transfiner is depicted in Figure~\ref{fig:model}. From the base object detection network,~\eg Mask R-CNN~\cite{he2017mask}, we employ a multi-scale deep feature pyramid. The object detection head then predicts bounding boxes as instance proposals. This component also generates a coarse initial mask prediction at low resolution. Given this input data, our aim is to predict highly accurate instance segmentation masks.


Since much of the segmentation errors are attributed to the loss of spatial resolution, we first define such \emph{incoherent regions} and analyze their properties in Section~\ref{sec:incoherent}. To identify and refine incoherent regions in multiple scales, we employ a quadtree, discussed in Section~\ref{quadtree_refinement}. The lightweight incoherent region detector takes as input the coarse initial mask alongside the multi-scale features, and predicts the incoherent regions for each scale in a cascaded manner. This allows ours Mask Transfiner to save huge computational and memory burdens, because only a small part of the high-resolution image features are processed by the refinement network itself.
Our refinement transformer, detailed in Section~\ref{sec:architecture}, operates in the detected incoherent regions. 
Since it operates on feature points on the constructed quadtree, and not in a uniform grid, we design a transformer architecture which jointly processes all incoherent nodes in all levels of the quadtree. 
Finally, we present the training strategy of Mask Transfiner along with the implementation details.



\begin{figure}[!t]
	\centering
	\includegraphics[width=1.0\linewidth]{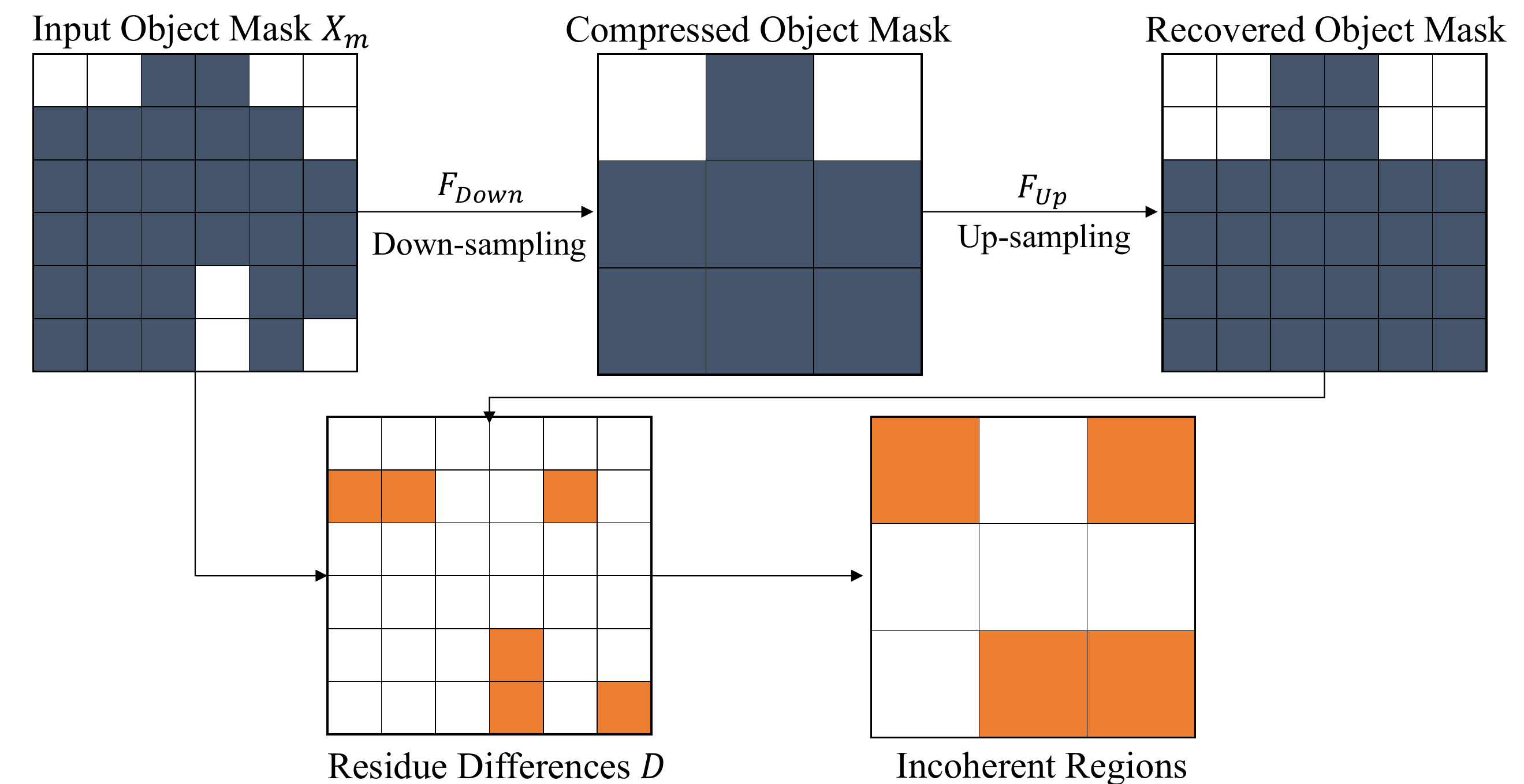}
	\caption{Illustration on incoherent regions definition by simulating mask information loss.}
	\label{fig:incoherent}
	\vspace{-0.2in}
\end{figure}

\subsection{Incoherent Regions}
\label{sec:incoherent}
Much of the segmentation errors produced by existing instance segmentation methods~\cite{he2017mask,dong2021solq} are due to the loss of spatial resolution, such as the mask downsampling operations, small RoI pooling size, and coefficients compression~\cite{dong2021solq, hu2021ISTR}, where mask prediction itself is performed at a coarse feature scale.
Despite its efficiency, low spatial resolution makes it challenging to predict accurate object boundaries, due to the loss of high-frequency details. In this section, we first define \emph{incoherent regions}, where mask information is lost due to reduced spatial resolution. Then, by analyzing their properties, we observe that a large portion of the errors are indeed located in these regions. 




\parsection{Definition of Incoherent Regions}
To identify incoherent regions, we simulate the loss of information due to downsampling in the network by also downsampling the mask itself. Specifically, information is lost in regions where the mask cannot be correctly reconstructed by a subsequent upsampling step,  as illustrated in Figure~\ref{fig:incoherent}. Formally, let $M_l$ be a binary ground-truth instance mask of an object at scale level $l$. The resolution at each scale level differs by a factor of 2, where $l=0$ is the finest and $l=L$ is the coarsest scale. We denote $2\times$ nearest neighbor down and upsampling by $\mathcal{S}_\downarrow$ and $\mathcal{S}_\uparrow$ respectively. The incoherent region at scale $l$ is then the binary mask achieved as, 
\begin{equation}
\label{eq1}
D_l = \mathcal{O}_\downarrow (M_{l-1} \oplus \mathcal{S}_\uparrow(\mathcal{S}_\downarrow(M_{l-1}))) \,. 
\end{equation}
Here, $\oplus$ denotes the logical `exclusive or' operation and $\mathcal{O}_\downarrow$ is $2\times$ downsampling by performing the logical `or' operation in each $2\times 2$ neighborhood. A pixel $(x, y)$ is thus incoherent $D_l(x,y)=1$ if the original mask value $M_{l-1}$ differs from its reconstruction in at least one pixel in the finer scale level.
%
%
Intuitively, incoherent regions are mostly strewn along object instance boundaries or high-frequency regions, consisting of points with missing or extra predicted wrong labels by coarse masks.  We provide the visualizations of them in Figure~\ref{fig:model} and Supp. file, which are sparsely and non-contiguously distributed on a typical image.
\vspace{-0.1in}

\begin{table}[!h]
	\caption{Experimental analysis of the incoherent regions on COCO \emph{val} set. Percent denotes the area ratio of incoherent regions in the object bounding boxes. Recall$_{\text{Err}}$ is the ratio for all wrongly predicted pixels per object. Acc is the accuracy rate for coarse mask predictions inside incoherent regions. AP$_{\text{Coarse}}$ is measured by using coarse mask predictions for whole object regions while AP$_{\text{GT}}$ only fills the incoherent regions with the ground truth labels.}
	\vspace{-0.1in}
	\centering
	\resizebox{0.75\linewidth}{!}{
		\begin{tabular}{c | c | c | c | c }
			\toprule
			Percent & Recall$_{\text{Err}}$ & Acc & AP$_{\text{GT}}$ & AP$_{\text{Coarse}}$  \\
			\midrule
			14\% & 43\% & 56\% & 51.0 & 35.5 \\
			\bottomrule
		\end{tabular}
	}
	\vspace{-0.1in}
	\label{tab:property}
\end{table}

\parsection{Properties of Incoherent Regions}
In Table~\ref{tab:property}, we provide an analysis of the incoherent regions defined above. It shows that a large portion of prediction errors are concentrated in these incoherent regions, occupying 43$\%$ of all wrongly predicted pixels, while only taking 14$\%$ to the corresponding bounding box areas. The accuracy of the coarse mask prediction in incoherent regions is 56$\%$. By fixing the bounding boxes detector, we conduct an oracle study to fill all these incoherent regions for each object with ground truth labels, while leaving the remaining parts as initial mask predictions. Compared to using initial mask predictions in the incoherent regions, the performance surges from 35.5 AP to 51.0 AP, indeed justifying they are critical for improving final performance.


\subsection{Quadtree for Mask Refinement}
\label{quadtree_refinement}

In this section, we describe our approach for detecting and refining incoherent regions in the image.
Our approach is based on the idea of iteratively detecting and dividing the incoherent regions in each feature scale. By only splitting the identified incoherent pixels for further refinement, our approach efficiently processes high-resolution features by only focusing on the important regions. To formalize our approach, we employ a quadtree structure to first identify incoherent regions across scales. We then predict the refined segmentation labels for all incoherent nodes in the quadtree, using our network detailed in Section~\ref{sec:architecture}. Finally, our quadtree is employed to fuse the new predictions from multiple scales by propagating the corrected mask probabilities from coarse to finer scales. 



\begin{figure*}[!t]
	\centering
    \includegraphics[width=0.95\linewidth]{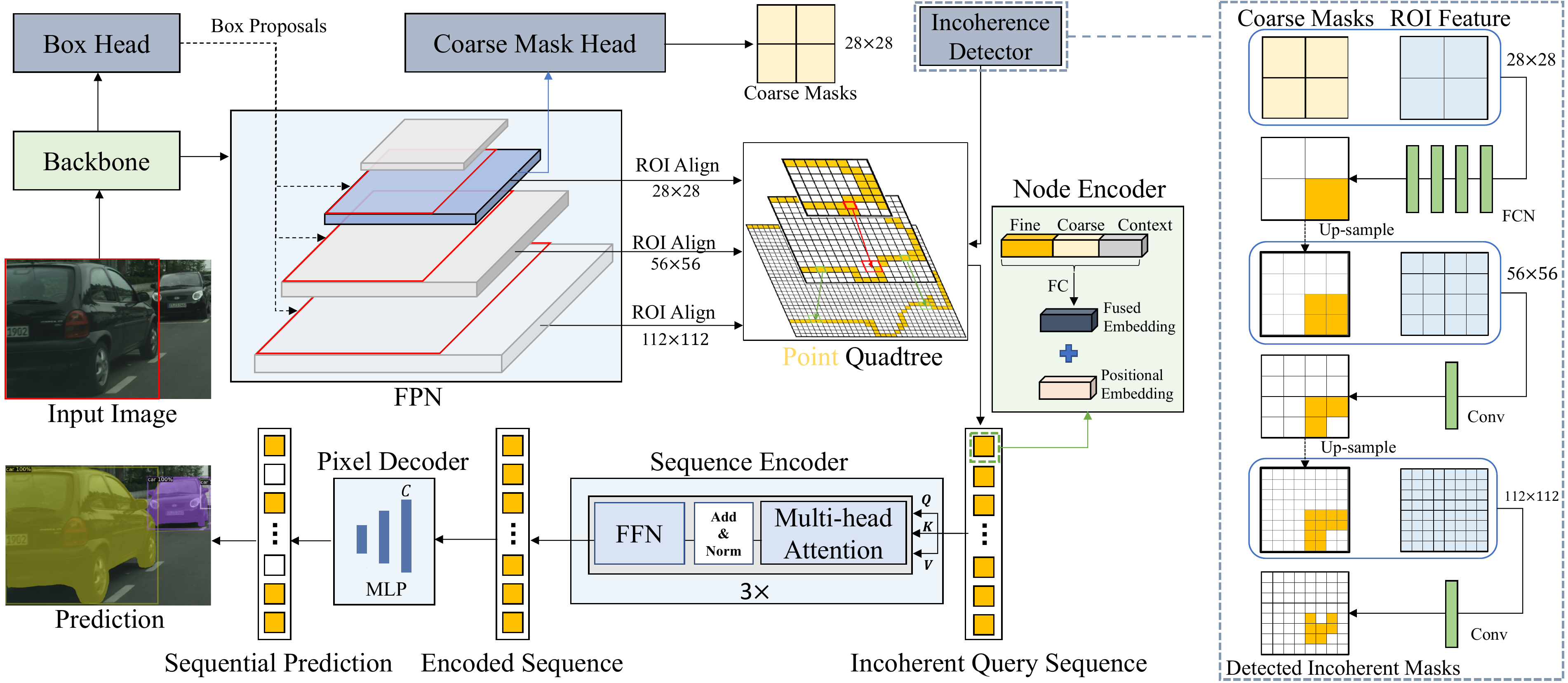}
	\vspace{-0.1in}
	\caption{The framework of Mask Transfiner. On the point quadtree, yellow point grids denote detected incoherent regions requiring further subdivision to four quadrants. The incoherent query sequence is composed of points across three levels of the quadtree for joint refinement. The encoder of Transfiner consists of node encoder and sequence encoder, while the pixel decoder is on top of each self-attended query pixel and output their final labels. The incoherence detector is detailed in the right part of the figure with detections on multi-level incoherent regions (Yellow). The higher-resolution detection is under the guidance of the predicted incoherent mask up-sampled from lower level.}
    \label{fig:model}
    \vspace{-0.25in}
\end{figure*}

\parsection{Detection of Incoherent Regions}
The right part of Figure~\ref{fig:model} depicts the design of our lightweight module to efficiently detect incoherent regions on a multi-scale feature pyramid. Following a cascaded design, we first concatenate the smallest features and coarse object mask predictions as input, and use a simple fully convolutional network (four 3$\times$3 Convs) followed by a binary classifier to predict the coarsest incoherence masks. Then, the detected lower-resolution masks are upsampled and fused with the larger-resolution feature in neighboring level to guide the finer incoherence predictions, where only single 1$\times$1 convolution layer is employed. During training, we enforce the groundtruth incoherent points in lower-level generated by Eq.~\ref{eq1} within the coverage of their parent points in higher-level feature map. 


\parsection{Quadtree Definition and Construction} We define a \emph{point quadtree} for decomposing the detected incoherent regions. Our structure is illustrated in Figure~\ref{fig:model}, where one yellow point in higher-level of FPN feature (such as feature resolution 28 $\times $28) has four quadrant points in its neighboring lower-level FPN feature map (such as resolution 56$\times$56). These are all feature points but with different granularities because they are on different pyramid levels. In contrast to the conventional quadtree `cells' used in computer graphics, where a quadtree `cell' can have multiple points, the subdivision unit for our point quadtree is always on a single point, with the division of points decided by the detected incoherent values and the threshold for the binary classifier.

Based on the detected incoherent points, we construct a multi-level hierarchical quadtree, beginning from using the detected points in the highest-level feature map as root nodes. These root nodes are selected for subdividing to their four quadrants on the lower-level feature map, with larger resolution and more local details. Note that at the fine level, only the quadrant points detected as incoherent could make a further break down and the expansion of incoherent tree nodes is restricted in regions corresponding to the incoherent predictions at the previous coarse level. 

\parsection{Quadtree Refinement}
We refine the mask predictions of the incoherent nodes of the quadtree using a transformer-based architecture. Our design is described in Sec.~\ref{sec:architecture}. It directly operates on the nodes of the quadtree, jointly providing refined mask probabilities at each incoherent node.

\parsection{Quadtree Propagation}
Given the refined mask predictions, we design a hierarchical mask propagation scheme that exploits our quadtree structure. Given the initial coarse masks predictions in low-resolution, Mask Transfiner first corrects the points labels belong to the root level of the quadtree, and then propagates these corrected point labels to their corresponding four quadrants in neighboring finer level by nearest neighbor interpolation. The process of labels correction is efficiently conducted on the incoherent nodes in a level-wise manner until reaching the finest quadtree level. Comparing to only correcting the labels of finest leaf nodes on the quadtree, it enlarges the refinement areas with negligible cost by propagating refinement labeled to leaf nodes of the intermediate tree levels. 

\subsection{Mask Transfiner Architecture}
\label{sec:architecture}



In this section, we describe the architecture of the refinement network, which takes as input the incoherent points on the built quadtree (Section~\ref{quadtree_refinement}) for final segmentation refinement. These points are sparsely distributed along the high-frequency regions across levels and not spatially contiguous. Thus, standard convolutional networks operating on uniform grids are not suitable. Instead, we design a refinement transformer, Mask Transfiner, that corrects the predictions of all incoherent quadtree nodes in parallel. 

Accurately segmenting ambiguous points requires both fine-grained deep features and coarse semantic information. The network therefore needs strong modeling power to sufficiently relate points and their surrounding context, including both spatial and cross-level neighboring points. Thus, a transformer, which can take sequential input and perform powerful local and non-local reasoning through the multi-head attention layers, is a natural choice for our Mask Transfiner design.
Compared to the MLP in~\cite{kirillov2020pointrend}, the strong global processing of transformers is a natural fit for our quadtree structure. It benefits the effective fusion of the multi-level feature points information with different granularities and the explicit modeling of pairwise point relations.


Figure~\ref{fig:model} shows the overall architecture of our Mask Transfiner. Based on the hierarchical FPN~\cite{lin2017feature}, instance segmentation is tackled in a multi-level and coarse-to-fine manner. Instead of using single-level FPN feature for each object~\cite{he2017mask}, Mask Transfiner takes as input sequence the sparsely detected feature points in incoherent image regions across the RoI feature pyramid levels, and outputs the corresponding segmentation labels.

\parsection{RoI Feature Pyramid} Given an input image, the CNN backbone network equipped with FPN first extracts hierarchical feature maps for downstream processing, where we utilize feature levels from $P_2$ to $P_5$. The base object detector~\cite{he2017mask, dong2021solq} predicts bounding boxes as instance proposals. Then the RoI feature pyramid is built by extracting RoI features across three different levels $\{P_i, P_{i-1}, P_{i-2}\}$ of FPN with increasing square sizes $\{28, 56, 112\}$. The starting level $i$ is computed as $i = \left \lfloor i_0 + \log_{2}(\sqrt{WH}/224) \right \rfloor$, where $i_{0} = 4$, $W$ and $H$ are the RoI width and height. The coarsest level features contain more contextual and semantic information, while the finer levels resolve more local details. 


\parsection{Input Node Sequence} 
Given the quadtree discussed in Section~\ref{quadtree_refinement} along with the associated FPN features for each node, we construct the input sequence for our transformer-based architecture. The sequence consists of all incoherent nodes from all three levels of the quadtree. The resulting sequence thus has a size of $C \times N$, where $N$ is the total number of nodes and $C$ is the feature channel dimension. Notably, $N \ll H W$ due to the high degree of sparsity. Moreover, the ordering of the sequence does not matter due to the permutation invariance of transformer. In contrast to standard transformer encoder, the encoder of Transfiner has two parts: the node encoder and the sequence encoder.

\parsection{Node Encoder} 
To enrich the incoherent points feature, the node encoder of Mask Transfiner encodes each quadtree node using the following four different information cues: \textbf{1)}~The fined-grained features extracted from corresponding location and level of the FPN pyramid. \textbf{2)}~The initial coarse mask prediction from the base detector provides region-specific and semantic information. 
\textbf{3)}~The relative positional encoding in each RoI encapsulates spatial distances and relations between nodes, capturing important local dependence and correlations. \textbf{4)}~The surrounding context for each node captures local details to enrich the information. For each node, we use features extracted from the 3$\times$3 neighborhood, compressed by a fully connected layer. Intuitively, this helps in localizing edges and boundaries, as well as capturing the local shape of the object. As illustrated in  Figure~\ref{fig:model}, the fine-grained features, coarse segmentation cues and context features are first concatenated and fused by a FC layer to original feature dimension. The positional embedding is then added to the resulting feature vector.

\parsection{Sequence Encoder and Pixel Decoder} 
Then, the sequence transformer encoder of Transfiner jointly processes the encoded nodes from all levels in the quadtree. The transformer thus performs both global spatial and inter-scale reasoning. Each sequence encoder layer has a standard transformer structure, formed by a multi-head self-attention module and a fully connected feed forward network (FFN). To equip the incoherent points sequence with adequate positive and negative references, we also use all feature points from the coarsest FPN level with small size 14$\times$14. Different from the standard transformer decoder~\cite{carion2020end} with deep attention layers, the pixel decoder in Mask Transfiner is a small two-layer MLP, which decodes the output query for each node in the tree, in order to predict the final mask labels.

\parsection{Training and inference}
Based on the constructed quadtree, we develop flexible and adaptive training and inference schemes for Mask Transfiner, where all detected incoherent nodes across quadtree levels are formed into a sequence for parallel prediction. During inference, to obtain final object masks, Mask Transfiner follows the quadtree propagation scheme (Section~\ref{quadtree_refinement}) after obtaining the refined labels for incoherent nodes. 
During training, the whole Mask Transfiner framework can be trained in an end-to-end manner. We employ a multi-task loss,
\vspace{-0.1in}
\begin{equation}
\mathcal{L} = \lambda_1\mathcal{L}_{\text{Detect}} +  \lambda_2\mathcal{L}_{\text{Coarse}} + \lambda_3\mathcal{L}_{\text{Refine}} + \lambda_4\mathcal{L}_{\text{Inc}} \,.
\end{equation}
Here,~$\mathcal{L}_{\text{Refine}}$ denotes the refinement with L$1$ loss between the predicted labels for incoherent nodes and their ground-truth labels. A Binary Cross Entropy loss $\mathcal{L}_{\text{Inc}}$ is for detecting incoherent regions. The detection loss $\mathcal{L}_{\text{Detect}}$ includes the localization and classification losses from the base detector, \eg Faster R-CNN~\cite{ren2015faster} or DETR detector. Finally,~$\mathcal{L}_{\text{Coarse}}$ represents the loss for the initial coarse segmentation prediction used by~\cite{he2017mask}. $\lambda_{\{1,2,3,4\}}$ are hyper-parameter weights $\{1.0, 1.0, 1.0, 0.5\}$. 

\parsection{Implementation Details}
Mask Transfiner is implemented on both the two-stage detector Faster R-CNN~\cite{ren2015faster} and query-based detector~\cite{carion2020end}. 
We design a 3-level quadtree and use the hyper-parameters and training schedules of Mask R-CNN implemented in Detectron2~\cite{wu2019detectron2} for the backbone and coarse mask head.
The Mask Transfiner encoder consists of three standard transformer layers. Each layer has four attention heads with feature dimension at 256. In our ablation study, R-50-FPN~\cite{he2016deep} and Faster R-CNN with 1$\times$ learning schedule are adopted.
For COCO leaderboard comparison, we adopt the scale-jitter with shorter image side randomly sampled from [640, 800], following training schedules in~\cite{lee2019centermask,chen2019tensormask,ke2021bcnet}.
More details are in the Supp. file.


\vspace{-0.1in}

\section{Experiments}

\subsection{Experimental Setup}

\parsection{COCO}
We perform experiments on COCO dataset~\cite{lin2014microsoft}, where we train our networks on 2017{\it train} and evaluate our results on both the 2017{\it val} and 2017{\it test-dev}. We employ the standard AP metrics and the recently proposed boundary IoU metrics~\cite{cheng2021boundary}. Notably,  AP$^{B}$ for boundary IoU is a measure focusing on boundary quality. Following~\cite{kirillov2020pointrend}, we also report AP$^\star$, which evaluates the \textit{val} set of COCO with significantly higher-quality LVIS annotations~\cite{gupta2019lvis} that can better reveal improvements in mask quality.

\parsection{Cityscapes} 
We report the results 
on Cityscapes~\cite{cordts2016cityscapes}, a high-quality instance segmentation dataset containing 2975, 500, 1525 images with resolution of 2048$\times$1024 for training, validation and test respectively. Cityscapes focus on self-driving scenes with 8 categories (\eg, car, person, bicycle). 

\parsection{BDD100K} We further train and evaluate Mask Transfiner on the BDD100K~\cite{bdd100k} instance segmentation dataset, which has 8 categories with 120K high-quality instance mask annotations. We follow the standard practice, using 7k, 1k, 2k images for training, validation and testing respectively.

\subsection{Ablation Experiments}

We conduct detailed ablation studies on the COCO validation set, analyzing the impact of the proposed incoherent regions and individual components of Mask Transfiner.

\parsection{Effect of the Incoherent Regions} 
Table~\ref{tab:property} presents
an analysis on the properties of incoherent regions described in Section~\ref{sec:incoherent}. It  reveals they are critical to the final segmentation performance.  Table~\ref{tab:effect_incoherent} presents analyzes the effectiveness of the detected incoherent regions by replacing the refinement regions with full RoIs or detected object boundary regions. 
Due to memory limitation, the full RoIs only uses output size 28$\times$28. The comparison shows the advantage of incoherent regions, with 1.8 AP and 0.7 AP gain over the use of 
full RoIs and detected boundary regions respectively.

To study the influence of incoherent regions on different pyramid levels, in Table~\ref{tab:effect_incoherent}, we also perform ablation experiments by removing the refinement regions of the Mask Transfiner in a level-wise order. We find that all three levels are beneficial to the final performance, while L$_1$ contributes most with 0.8 AP increase, where L$_1$ denotes the root level of Mask Transfiner with the smallest feature size.

\begin{table}[!t]
    \vspace{-0.1in}
	\caption{Effect of the incoherent regions on COCO \textit{val} set. AP$^B$ is evaluated Boundary IoU~\cite{cheng2021boundary} while AP$^\star$ uses LVIS annotations.}
	\vspace{-0.1in}
	\centering
	\resizebox{0.8\linewidth}{!}{
		\begin{tabular}{c | c | c | c | c}
			\toprule
			Region Type & AP & AP$^B$ & AP$^{\star}$ & AP$^{\star}_{50}$  \\
			\midrule
			Full RoIs (28 $\times$ 28) & 35.5 & 21.4 & 38.3 & 59.5 \\
			Boundary regions & 36.6 & 23.8 & 40.1 & 60.2 \\
			Incoherent regions & \textbf{37.3} & \textbf{24.2} & \textbf{40.5} & \textbf{60.7} \\
			\midrule
			Incoherent regions (\textbf{w/o} L$_1$) & 36.5 & 23.5 & 39.8 & 59.7 \\
			Incoherent regions (\textbf{w/o} L$_2$) & 36.8 & 23.8 & 40.2 & 60.1 \\
			Incoherent regions (\textbf{w/o} L$_3$) & 36.7 & 23.6 & 40.0 & 59.9 \\
			\bottomrule
		\end{tabular}
	}
	\vspace{-0.15in}
	\label{tab:effect_incoherent}
\end{table}

\begin{table}[!t]
	\caption{Effect of lower-level masks guidance in detecting incoherent regions on COCO \textit{val}. AP and AP$^B$ are final performance.}
	\vspace{-0.1in}
	\centering
	\resizebox{0.8\linewidth}{!}{
		\begin{tabular}{c | c | c | c | c }
			\toprule
		    Lower-level Guidance & Acc & Recall & AP & AP$^B$   \\
			\midrule
			 & 79$\%$ & 73$\%$ & 36.6 & 23.7 \\
			\checkmark & \textbf{84$\%$} & \textbf{86$\%$} & \textbf{37.3} & \textbf{24.2} \\
			\bottomrule
		\end{tabular}
	}
	\vspace{-0.12in}
	\label{tab:guidance}
\end{table}

\begin{table}[!t]
	\caption{Analysis of node encoding cues on COCO \textit{val} set.}
	\vspace{-0.1in}
	\centering
	\resizebox{0.8\linewidth}{!}{
		\begin{tabular}{c | c | c | c | c | c | c | c }
			\toprule
			Fine & Coarse & Pos. & Context & AP & AP$^B$ & AP$^{\star}$ & AP$^{\star}_{50}$\\
			\midrule
			\checkmark & & &  & 33.8 & 20.1 & 37.0 & 53.8 \\
			\checkmark & \checkmark & & & 34.2 & 20.4 & 37.3 & 54.3 \\
			\checkmark & \checkmark & \checkmark & & 36.8 & 23.9 & 40.1 & 60.1 \\
			\checkmark & \checkmark & \checkmark  & \checkmark  & \textbf{37.3} & \textbf{24.2} & \textbf{40.5} & \textbf{60.7} \\
			\bottomrule
		\end{tabular}
	}
	\vspace{-0.25in}
	\label{tab:component}
\end{table}

\begin{table*}[!t]
     \vspace{-0.3in}
	\begin{minipage}[t]{0.35\linewidth}
		\caption{Analysis of the quadtree depth on the COCO \textit{val} using R50-FPN as backbone.}
		\centering
		\vspace{-0.1in}
		\resizebox{1.0\linewidth}{!}{
			\begin{tabular}{c | c | c | c | c | c | c | c}
				\toprule
				Depth & Output size & AP & AP$^{\star}$ & AP$_{L}$ & AP$_{M}$ & AP$_{S}$ & FPS \\
				\midrule
				0 & 28$\times$28 & 35.2 & 37.6 & 50.3 & 37.7  & 17.2 & 12.3 \\
				\midrule
				1 & 28$\times$28 & 35.5 & 38.4 & 50.9 & 38.1  & 17.2 & 10.6 \\
				2 & 56$\times$56 & 36.2 & 39.1 & 51.9 & 38.7  & 17.3  & 8.9 \\
			    3 & 112$\times$112 & \textbf{37.3} & 40.5 & 52.9 & \textbf{39.5}  & \textbf{17.5} & 7.1 \\
			    4 & 224$\times$224 & 37.1 & \textbf{40.7} & \textbf{53.1} & 39.3  & 17.4 & 5.2 \\
				\bottomrule
			\end{tabular}
		}
		\label{tab:stage}
	\end{minipage}
	\hfill
	\begin{minipage}[t]{0.32\linewidth}
		\caption{Mask Transfiner vs.\ MLP and CNN on COCO \textit{val} set using ResNet-50-FPN.}
		\centering
		\vspace{-0.1in}
		\resizebox{1.0\linewidth}{!}{
			\begin{tabular}{l | c | c | c| c}
				\toprule
				Model & AP & AP$^B$ & AP$^{\star}$ & AP$^{\star}_{50}$ \\
				\midrule
				CNN (full regions, 56 $\times$ 56) & 35.7 & 21.8 & 38.7 & 58.8 \\
				\midrule
				MLP (full regions, 56 $\times$ 56) & 36.1 & 23.4 & 39.2 & 59.2 \\
				MLP (PointRend~\cite{kirillov2020pointrend}, 112 $\times$ 112) & 36.2 & 23.1 & 39.1 & 59.0 \\
				MLP (incoherent regions) & 36.4 & 23.7 & 39.7 & 59.8 \\
				\midrule
				Mask Transfiner (D = 3, H = 4) & 37.3 & 24.2 & 40.5 & 60.7 \\
				Mask Transfiner (D = 3, H = 8) & 37.1 & 24.1 & 40.2 & 60.8 \\
				Mask Transfiner (D = 6, H = 4) & 37.4 & 24.4 & 40.6 & 60.9 \\
				\bottomrule
			\end{tabular}
		}
		\label{tab:struct_compare}
	\end{minipage}
	\hfill
	\begin{minipage}[t]{0.32\linewidth}
		\caption{Efficacy of Transfiner compared to standard attention models on COCO \textit{val}. NLA denotes non-local attention~\cite{wang2018non}.}
		\centering
		\vspace{-0.1in}
		\resizebox{1.0\linewidth}{!}{
			\begin{tabular}{l | c | c | c | c}
				\toprule
				Model & AP & FLOPs (G) & Memory (M) & FPS\\
				\midrule
				NLA~\cite{wang2018non} (112$\times$112) & 36.3 & 24.6 & 8347 & 4.6 \\
				NLA~\cite{wang2018non} (224$\times$224) & 36.6 & 80.2 & 18091 & 2.4 \\
		        \midrule
				Transformer~\cite{carion2020end} (28$\times$28) & 36.1 & 37.2 & 4368 & 6.9 \\
				Transformer~\cite{carion2020end} (56$\times$56) & 36.5 & 68.3 & 17359 & 2.1 \\
				\midrule
				Mask Transfiner (112$\times$112) & \textbf{37.3} & \textbf{16.8} & \textbf{2316} & \textbf{7.1} \\
				Mask Transfiner (224$\times$224) & 37.1 & 38.1 & 4871 & 5.2 \\
				\bottomrule
			\end{tabular}
		}
		\label{tab:trans_compare}
	\end{minipage}
	\vspace{-0.2in}
\end{table*}

\parsection{Ablation on the Incoherent Regions Detector} 
We evaluate the performance of the light-weight incoherent region detector by computing its recall and accuracy rates. In Table~\ref{tab:guidance}, with the guidance of the predicted incoherent mask up-sampled from lower level (Figure~\ref{fig:model}), the recall rate of detected incoherent regions has an obvious improvement from 74$\%$ to 86$\%$, and the accuracy rate also increases from 79$\%$ to 84$\%$. Note that recall rate is more important here to cover all the error-prone regions for further refinements.

\parsection{Effect of Incoherent Points Encoding} 
We analyze the effect of the four information cues in the incoherent points encoding. In Table~\ref{tab:component}, comparing to only using the fine-grained feature, the coarse segmentation features with semantic information brings a gain of 0.4 point AP. The positional encoding feature has a large influence on model performance by significantly improving 2.6 points on AP and 3.5 points on AP$^B$ respectively. The positional encoding for incoherent points are crucial, because transformer architecture is permutation-invariant and the segmentation task is position-sensitive. The surrounding context feature further promotes the segmentation results from 36.8 AP to 37.3 AP by aggregating local neighboring details.

\begin{figure}[!t]
	\centering
	\includegraphics[width=1.0\linewidth]{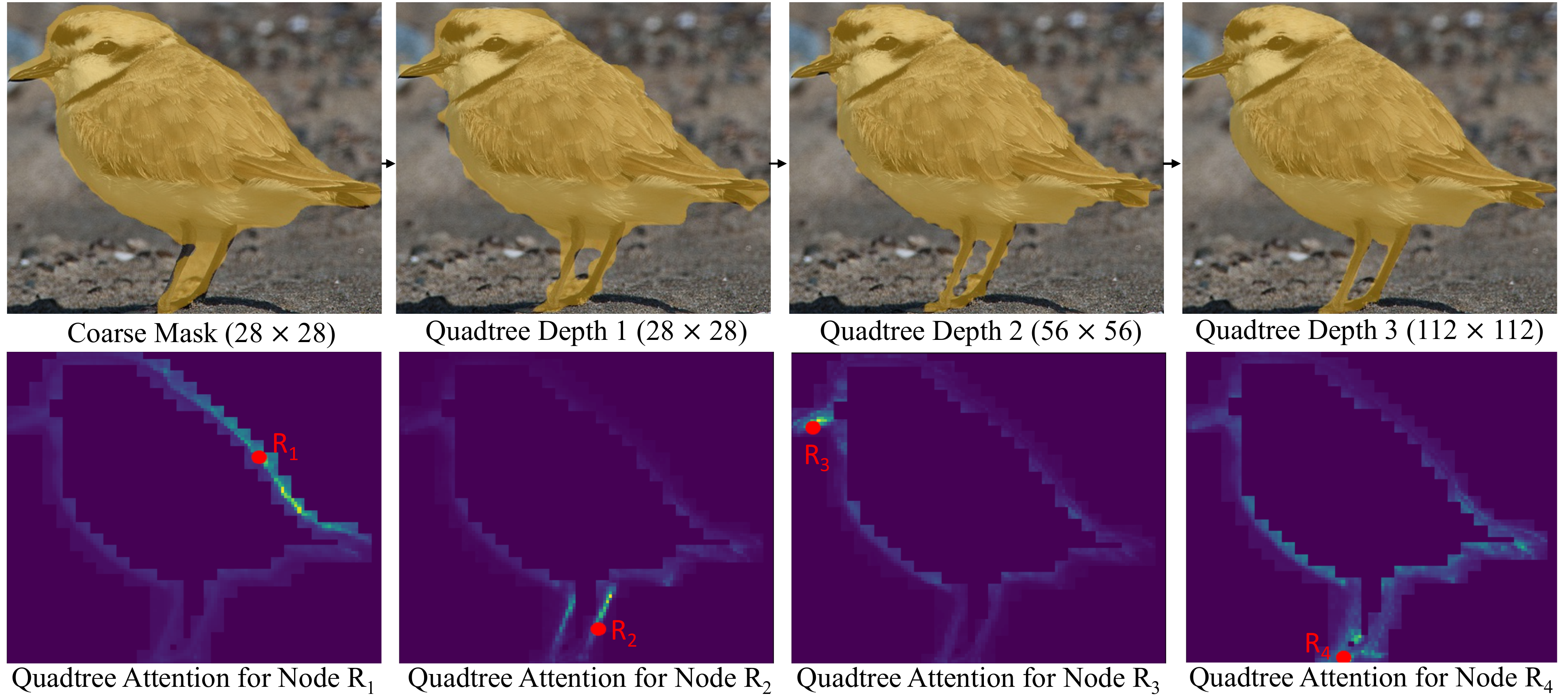}
	\vspace{-0.27in}
	\caption{Qualitative results comparison between the coarse mask predictions by our baseline~\cite{he2016deep} and the refinement results with various depths of the quadtree built on detected incoherent regions. The bottom row visualizes the quadtree attention weights distribution in the sparse incoherent regions for four sampled red nodes.}
	\label{fig:depth_ab}
	\vspace{-0.25in}
\end{figure}

\parsection{Influence of Quadtree Depths} 
In Table~\ref{tab:stage}, we study the influence on hierarchical refinement stages by constructing the quadtree in our Mask Transfiner with different depths. Depth 0 denotes the baseline using coarse head mask prediction \textbf{w/o} refinement steps. The output size grows twice larger than its preceding stage. By varying the output sizes from 28$\times$28 to 224$\times$224, the mask AP$^\star$ increases from 38.4 to 40.7 with increased tree depth. This reveals that models with more levels and larger output sizes for an object indeed brings more gain to segmentation performance. The large objects benefit most from the increasing sizes with an improvement of 2.8 point in AP$_L$. We further find that the performance saturates when the output size is larger than 112$\times$112, while the 3-stage Transfiner also has a lower computational cost and runs at 7.1 fps. Figure~\ref{fig:depth_ab} visualizes results with increasing quadtree depths, where masks become substantially finer detail around object boundaries.

\parsection{Mask Transfiner vs.\ MLP and CNN}
We compare different popular choices of the refinement networks, including the MLP and CNN structures. MLP is implemented with three hidden layers of 256 channels~\cite{kirillov2020pointrend}, while CNN is a FCN with four convolution layers with 3$\times$3 kernels~\cite{he2017mask}.
Note that for full refinement regions, CNN and MLP are limited to the RoI size 56 $\times$ 56 due to memory limitations, and CNN is not suitable for incoherent regions because uniform grids are required.
In Table~\ref{tab:struct_compare}, our Mask Transfiner outperforms the MLP by 0.9 AP, benefiting from the non-local pixel-wise relation modeling, where we use the same incoherent regions on all three quadtree levels for fair comparison.
Moreover, we investigate the influence of layer depth $D$ and width $W$ of Mask Transfiner and find that deeper and wider attention layers only lead to minor performance change.
In Figure~\ref{fig:depth_ab}, we visualize the sparse quadtree attention maps of the last sequence encoder layer of the Transfiner, focusing on a few incoherent points. The encoder already seems to distinguish between foreground instances and background, where the neighboring attended regions of point R$_{1}$ are separated by the object boundary.

\begin{figure*}[!t]
	\centering
    \includegraphics[width=1.0\linewidth]{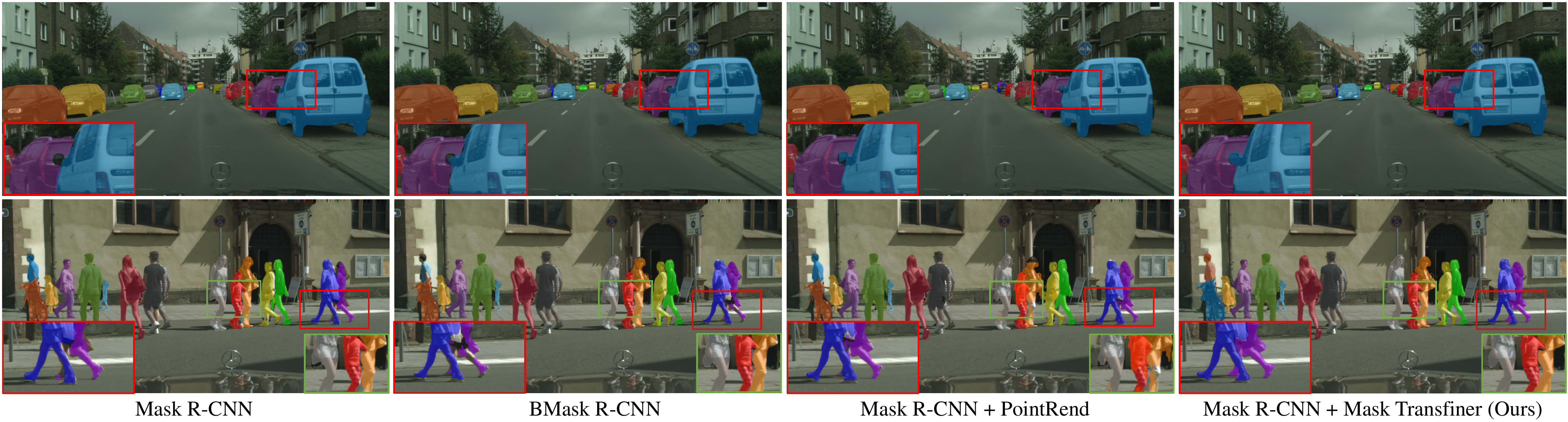}
	\vspace{-0.3in}
	\caption{Qualitative comparisons with instance segmentation methods Mask R-CNN~\cite{he2017mask}, BMask R-CNN~\cite{ChengWHL20}, PointRend~\cite{kirillov2020pointrend} and our Mask Transfiner on Cityscapes \textit{val} set. Mask Transfiner produces more natural boundaries while revealing details for small parts, such as the rear mirrors of the car and the high-heeled shoes. Zoom in for better view. Refer to the supplemental file for more visual comparisons.}
    \label{fig:quali_comparison}
    \vspace{-0.1in}
\end{figure*}

\parsection{Efficacy of Quadtree Structure} 
Table~\ref{tab:trans_compare} compares Mask Transfiner with different attention mechanisms. Compared to pixels relation modeling using 3-layer non-local attention~\cite{wang2018non} or standard transformer~\cite{carion2020end,vaswani2017attention}, Mask Transfiner not only obtains higher accuracy but also is very efficient in computation and memory consumption. For example, Mask Transfiner with multi-head attention uses 3 times less memory than the non-local attention given  same output size, due to the small number of incoherent pixels. Compared to standard transformer operating on full RoI regions of much smaller size 56$\times$56, the quadtree subdivision and inference allows Mask Transfiner to produce a high-resolution 224$\times$224 prediction using only half of the FLOPs computation. Note that the standard transformer with output size 112$\times$112 runs out of memory in our experiments. 


\parsection{Effect of Multi-level Joint Refinement}
Given incoherent nodes from the 3-level quadtree, Transfiner forms all of them into a sequence for joint refinement in single forward pass. In Table~\ref{tab:prop}, we compare it with separately refining the quadtree nodes on each level with multiple sequences. The performance boost of 0.6 AP$^{\star}$ shows the benefit of multi-scale feature fusion and richer context in global reasoning.

\parsection{Effect of Quadtree Mask Propagation}
During inference, after Mask Transfiner has refined all incoherent points, we utilize a hierarchical coarse-to-fine mask propagation scheme along the quadtree levels to obtain the final predictions. Comparing to only correcting the labels of finest leaf nodes on the quadtree in Table~\ref{tab:prop}, the propagation enlarges the refinement areas and improves the performance from 36.5 AP to 37.0 AP. The propagation brings negligible computation because the new labels for the quadrant leaf (coherent) nodes in intermediate tree levels are obtained via duplicating the refined label values of their parents.
\vspace{-0.05in}

\begin{table}[!h]
	\caption{Effect of the multi-level joint refinement (MJR) and quadtree mask propagation (QMP) on COCO \textit{val} set.}
	\vspace{-0.1in}
	\centering
	\resizebox{0.7\linewidth}{!}{
		\begin{tabular}{c | c | c | c | c | c}
			\toprule
		    MJR & QMP & AP & AP$^B$ & AP$^{\star}$ & AP$^{\star}_{50}$  \\
			\midrule
			 &  & 36.5 & 23.7 & 39.6 & 59.7 \\
			 \checkmark &  & 36.9 & 23.9 & 40.2 & 60.2 \\
			 & \checkmark & 37.0 & 24.0 & 40.1 & 60.2 \\
			\checkmark & \checkmark & \textbf{37.3} & \textbf{24.2} & \textbf{40.5} & \textbf{60.7} \\
			\bottomrule
		\end{tabular}
	}
	\vspace{-0.1in}
	\label{tab:prop}
\end{table}


\begin{table*}[!t]
	\caption{Comparison with SOTA methods on COCO {\it test-dev} and \textit{val} set. All methods are trained on COCO \emph{train2017}. $\dagger$: trained with DCN~\cite{zhu2019deformable}. AP$^{\star}$ denotes evaluation using LVIS~\cite{gupta2019lvis} annotation and AP$^B$ denotes using Boundary IoU~\cite{cheng2021boundary}.}
	\vspace{-0.25in}
	\begin{center}{\small
			\resizebox{0.8\linewidth}{!}{
				\begin{tabular}{l|c|c|c|cc|c|ccc}
					\hline
					Method & Backbone & Type & AP & AP$^{\star}_{\text{val}}$ & AP$^B_{\text{val}}$ & AP$^{\text{Box}}$ & AP$_{S}$ & AP$_{M}$ & AP$_{L}$ \\ 
					\hline
					Mask R-CNN~\cite{he2017mask} & R50-FPN & Two-stage & 37.5 & 38.2 & 21.2 & 41.3 & 21.1 & 39.6 & 48.3 \\ 
					PointRend~\cite{kirillov2020pointrend} & R50-FPN & Two-stage & 38.1 & 39.7 & 23.5 & 41.5 & 18.8 & 40.2 & 49.4 \\
					B-MRCNN~\cite{ChengWHL20} & R50-FPN & Two-stage & 37.8 & 39.8 & 23.5 & 41.6 & 19.7 & 40.3 & 49.6 \\
					BPR~\cite{tang2021look} & R50-FPN & Two-stage & 38.4 & 40.2 & 24.3 & 41.3 & 20.2 & 40.5 & 49.7 \\
					Mask Transfiner & R50-FPN & Two-stage & 39.4 & 42.3 & 26.0 & 41.8 & 22.3 & 41.2 & 50.2 \\
					\textbf{Mask Transfiner$^\dagger$} & R50-FPN & Two-stage & \textbf{40.5} & \textbf{43.1} & \textbf{26.8} & \textbf{43.2} & \textbf{22.8} & \textbf{42.3} & \textbf{52.5} \\
					\hline
					Mask R-CNN~\cite{he2017mask} & R101-FPN & Two-stage & 38.8 & 39.3 &  23.1 & 43.1 & 21.8 & 41.4 & 50.5 \\ 
					PointRend~\cite{kirillov2020pointrend} & R101-FPN & Two-stage & 39.6 & 41.4 & 25.3 & 43.3 & 19.8 & 42.6 & 53.7 \\
				    MS R-CNN$^\dagger$~\cite{huang2019mask} & R101-FPN & Two-stage & 39.6 & 41.1 & 25.0 & 44.1 & 18.9 & 42.7 & 55.1 \\
					HTC~\cite{chen2019hybrid} & R101-FPN & Two-stage & 39.7 & 42.5 & 25.4 & \textbf{45.9} & 21.0 & 42.2 & 53.5 \\
					RefineMask~\cite{refinemask} & R101-FPN & Two-stage & 39.4 & 42.3 & 26.8 & 43.8 & 21.6 & 42.0 & 53.1 \\
					BCNet~\cite{ke2021bcnet} & R101-FPN & Two-stage & 39.8 & 41.9 & 26.1 & 43.5 & 22.7 & 42.4 & 51.1 \\
					Mask Transfiner & R101-FPN & Two-stage & 40.7 & 43.6 & 27.3 & 43.9 & 23.1 & 42.8 & 53.8 \\
					\textbf{Mask Transfiner$^\dagger$} & R101-FPN & Two-stage & \textbf{42.2} & \textbf{45.0} & \textbf{28.6} & 45.8 & \textbf{24.1} & \textbf{44.8} & \textbf{55.4} \\
					\hline
					\hline
					ISTR~\cite{hu2021ISTR}  & R50-FPN & Query-based & 38.6 & 39.5 & 23.0 & 46.8 & 22.1 & 40.4 & 50.6 \\
					QueryInst~\cite{QueryInst} & R50-FPN & Query-based & 39.9 & 42.1 & 25.1 & 44.5 & 22.9 & 41.7 & 51.9 \\
					SOLQ~\cite{dong2021solq} & R50-FPN & Query-based & 39.7 & 39.8 & 23.3 & \textbf{47.8} & 21.5 & 42.5 & 53.1 \\ 
					\textbf{Mask Transfiner} & R50-FPN & Query-based & \textbf{41.6} & \textbf{45.4} & \textbf{28.2} & 46.5 & \textbf{24.2} & \textbf{44.6} & \textbf{55.2} \\
					\hline
		\end{tabular}}}
	\end{center}
	\label{table:fully}
	\vspace{-0.3in}
\end{table*}

\begin{table}[!h]
	\caption{Performance comparison between two-stage instance segmentation methods on Cityscapes \textit{val} set using R50-FPN.}
	\vspace{-0.1in}
	\centering
	\resizebox{0.9\linewidth}{!}{
		\begin{tabular}{l | c | c | c | c }
			\toprule
			Method & AP$^{B}$ & AP$^{B}_{50}$ & AP & AP$_{50}$ \\
			\midrule
			Mask R-CNN (Baseline)~\cite{he2017mask} & 11.4 & 37.4 & 33.8 & 61.5 \\
			PointRend~\cite{kirillov2020pointrend} & 16.7 & 47.2 & 35.9 & 61.8 \\
			BMask R-CNN~\cite{ChengWHL20} & 15.7 & 46.2 & 36.2 & 62.6 \\
			Panoptic-DeepLab~\cite{cheng2020panoptic} & 16.5 & 47.7 & 35.3 & 57.9  \\
			RefineMask~\cite{refinemask} & 17.4 & 49.2 & 37.6 & 63.3  \\ 
			\midrule
			Mask Transfiner (Ours) & \textbf{18.0} & \textbf{49.8} & \textbf{37.9} & \textbf{64.1}  \\
			\bottomrule
		\end{tabular}
	}
	\vspace{-0.15in}
	\label{tab:cityscape_comp}
\end{table}

\begin{table}[!h]
	\caption{Performance comparison between instance segmentation methods on BDD100K \textit{val} set.}
	\vspace{-0.1in}
	\centering
	\resizebox{0.9\linewidth}{!}{
		\begin{tabular}{l | c | c | c }
			\toprule
			Method & Backbone & AP$_{\text{mask}}$ & AP$_{\text{box}}$ \\
			\midrule
			Mask R-CNN (Baseline)~\cite{he2017mask} & R101-FPN & 20.5 & 26.1 \\
			Cascade Mask R-CNN~\cite{cai2019cascade} & R101-FPN & 19.8 & 24.7 \\
			Mask R-CNN + DCNv2~\cite{zhu2019deformable} & R101-FPN & 20.9 & 26.0 \\
			HRNet~\cite{wang2020deep} & HRNet-w32 & 22.5 & \textbf{28.2} \\
			\midrule
			Mask Transfiner (Ours) & R101-FPN & \textbf{23.6} & 26.2 \\
			\bottomrule
		\end{tabular}
	}
	\vspace{-0.25in}
	\label{tab:bdd_comp}
\end{table}


\subsection{Comparison with State-of-the-art}
We compare our approach with the state-of-the-art methods on the benchmarks COCO, Cityscapes and BDD100K, where Mask Transfiner outperforms all existing methods without bells and whistles, demonstrating efficacy on both two-stage and query-based segmentation frameworks. Codes and models will be released upon publication.


\parsection{COCO} Table~\ref{table:fully} compares Mask Transfiner with state-of-the-art instance segmentation methods on COCO dataset.
Transfiner achieves consistent improvement on different backbones and object detectors, demonstrating its effectiveness by outperforming RefineMask~\cite{refinemask} and BCNet~\cite{ke2021bcnet} by 1.3 AP and 0.9 AP using R101-FPN and Faster R-CNN, and exceeding QueryInst~\cite{QueryInst} by 1.7 AP using query-based detector~\cite{carion2020end}. Note QueryInst consists of six-stage refinement in parallel with far more parameters to optimize.
Besides, we find that Transfiner using Faster R-CNN and R50-FPN with much lower object detection performance still achieves comparable segmentation results with query-based methods~\cite{dong2021solq,hu2021ISTR} on mask AP, and over 2 points gain in boundary AP$^B$, further validating the higher AP achieved by Transfiner is indeed contributed by the fine-grained masks.

\parsection{Cityscapes}
The results of Cityscapes benchmark is tabulated in Table~\ref{tab:cityscape_comp}, where Mask Transfiner achieves the best mask AP 37.6 and boundary AP$^{B}$ 18.0. Our approach significantly surpasses existing SOTA methods, including PointRend~\cite{kirillov2020pointrend}
and BMask R-CNN~\cite{ChengWHL20} by a margin of 1.3 AP$^{B}$ and 2.3 AP$^{B}$ using the same Faster R-CNN detector. Compared to our baseline Mask R-CNN~\cite{he2017mask}, Transfiner greatly improves the boundary AP from 11.4 to 18.0, which shows the effectiveness of the quadtree refinement. 

\parsection{BDD100K}
Table~\ref{tab:bdd_comp} shows results on BDD100K dataset, where Mask Transfiner obtains the highest AP$_{\text{mask}}$ of 23.5 and outperforms the baseline~\cite{he2016deep} by 3 points under the comparable AP$_{\text{Box}}$. The significant advancements reveals the high accuracy of the predicted masks by Transfiner.



\parsection{Qualitative Results}
Figure~\ref{fig:quali_comparison} shows qualitative comparisons on Cityscapes, where our Mask Transfiner produces masks with substantially
higher precision and quality than previous methods~\cite{he2017mask, ChengWHL20, kirillov2020pointrend}, especially for the hard regions, such as the small rear mirrors and high-heeled shoes. Refer to supplementary file for more visual comparisons.

\section{Conclusion}
We present Mask Transfiner, a new high-quality and efficient instance segmentation method.  Transfiner first detects and decomposes the image regions to build a hierarchical quadtree. Then, all points on the quadtree are transformed into to a query sequence for our transformer to predict final labels. In contrast to previous segmentation methods using convolutions limited by uniform image grids, Mask Transfiner produces high-quality masks with low computation and memory cost. We validate the efficacy of Transfiner on both the two-stage and query-based segmentation frameworks, and show that Transfiner achieves large performance advantages on COCO, Cityscapes and BDD100K. 
A current limitation is the fully supervised training required by our Mask Transfiner as well as the competing methods. Future work will strive towards relaxing this assumption.

\section{Appendix}
We first provide more implementation and training/inference details of Mask Transfiner on three instance segmentation benchmarks (Section~\ref{sec:details}). Then we conduct more experimental analysis and discussion of comparison between Mask Transfiner and other methods (Section~\ref{sec:supp_exp}).
We further present more qualitative results comparisons on COCO~\cite{lin2014microsoft}, BDD100K~\cite{bdd100k} and Cityscapes~\cite{cordts2016cityscapes} datasets in various scenes (Section~\ref{sec:qual_comp}).
Finally, we visualize quadtree attention weights, detected incoherent regions and segmentation results with various quadtree depths (Section~\ref{sec:vis_analysis}), including failure cases analysis.

\subsection{More Implementation Details}
\label{sec:details}

\paragraph{Implementation and Training/Inference Details} We implement Mask Transfiner based on Detectron2~\cite{wu2019detectron2}, where SGD is used with 0.9 momentum and 1K constant warm-up iterations. The weight decay is set to 0.0001. On the two-stage and query-based frameworks, we employ Mask Transfiner using Faster R-CNN~\cite{ren2015faster} and DETR~\cite{carion2020end} detectors respectively while leaving the RoI pyramid construction and refinement transformer unchanged. 

To make the detection on incoherent regions more robust, we adopt jittering operations along the boundaries of the ground truth incoherent regions because in our case, the recall rate of the detection to cover all the incoherent regions play a more critical role in influencing final performance. We use 0.5 as the threshold for the binary incoherence classifier. For the experiment of Table 2 in the paper, the boundary regions are pixels within two-pixel Euclidean distance to the detected object mask contours on all three levels of the object feature pyramid, where the object boundary detector~\cite{ke2021bcnet} is used. The coarse mask head is composed of a FCN network with four 3×3 Convs attached on the ROI feature of size 28$\times$28. 

During training, we randomly permute the order of the incoherent points for each object and select 300 of them (100 per quadtree level), so as to maintain the same sequence length for each object for batch efficiency. We adopt the horizontal flipping and scale data augmentation during training following~\cite{kirillov2020pointrend}. 

\begin{figure*}[!t]
	\centering
	\vspace{-0.18in}
	\includegraphics[width=1.0\linewidth]{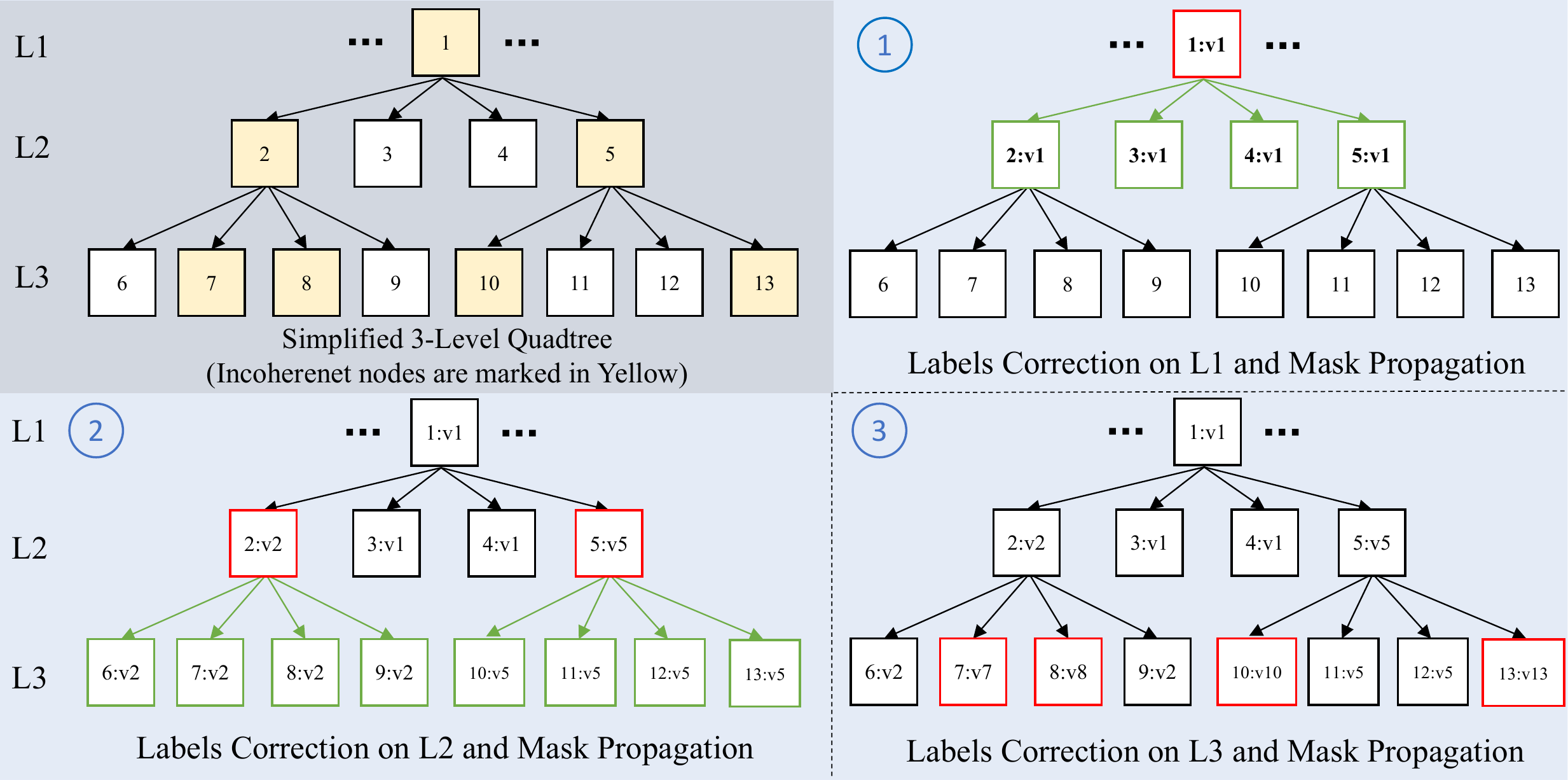}
	\vspace{-0.2in}
	\caption{The simplified illustration of mask propagation on a 3-level quadtree during inference. Given the detected incoherent nodes and their refined predictions, in Step 1, Mask Transfiner corrects the nodes labels belonging to L1 level of the quadtree, and then propagates these corrected labels to their corresponding four quadrants in L2 level. In Step 2, the process of labels correction is efficiently conducted on the incoherent nodes in L2 and further propagating to L3. This process is recursive until reaching the finest quadtree level.}
	\label{fig:mask_prop}
	\vspace{-0.2in}
\end{figure*}

During inference, no test-time augmentation is used. We employ a hierarchical propagation scheme based on the quadtree structure from coarse to finer scales (detailed in Section 3.1 of the paper) and the refined incoherent nodes predictions. In Figure~\ref{fig:mask_prop}, we further illustrate the mask propagation process with a simplified 3-level quadtree. The incoherent nodes number $N$ with their refined predictions value $V_n$ are formatted in $N:V_n$, where $\{1:\text{v}1, 2:\text{v}2, 5:\text{v}5, 7:\text{v}7, 8:\text{v}8, 10:\text{v}10, 13:\text{v}13\}$ are incoherent nodes numbers and prediction values pairs in our given example. We break down the mask correction and propagation into 3 steps corresponding to 3 levels of the quadtree with visualizations. Comparing to only correcting the labels of finest leaf nodes on the quadtree, it enlarges the refinement areas with negligible cost by propagating refinement labeled to leaf nodes $\{3, 4, 6, 9, 11, 12\}$. We validate the effect of quadtree mask propagation in Table 8 of the paper.

\vspace{-0.15in}
\paragraph{COCO:} We set 16 images per mini-batch. Following~\cite{kirillov2020pointrend}, our training schedule is 60k / 20k / 10k with updating learning rates 0.02 / 0.002 / 0.0002 respectively.
For ablation study, our method is trained on four GPUs using ResNet-50, where we use SGD for optimization and set initial learning rate to 0.01 with total batch size 8. We train Mask Transfiner for 12 epochs (taking about 8 hours with NVIDIA RTX 2080 Ti), and decrease the learning rate by 0.1 after 8 and 11 epochs. 

\vspace{-0.15in}
\paragraph{Cityscapes:} We adopt 8 images per mini-batch and the training schedule is 18k / 6k updates at learning rates of 0.01 / 0.001 respectively. During training, the images are resized randomly to a shorter edge from [800, 1024] pixels with a step of 32 pixels. The inference images are resized to a shorter edge size of 1024 pixels. For Cityscapes evaluation, we train the models on the fine annotations of the train set with 64 epochs following~\cite{kirillov2020pointrend,ChengWHL20}. 

\vspace{-0.15in}
\paragraph{BDD100K:}
We use 16 images per mini-batch and and the training schedule is 22k / 4k/ 4k updates at learning rates of 0.02 / 0.002/ 0.0002 respectively. During training, the images are resized randomly to a shorter edge from [600, 720] pixels with a step of 24 pixels. During inference, the images are resized to a shorter edge size of 720 pixels. Note all compared methods are trained with the same schedules and image size settings.

\subsection{More Experimental Analysis}
\label{sec:supp_exp}

\paragraph{Accuracy Comparison} In Table 9 of the main text, we compare the accuracy of Mask Transfiner with previous methods and find that Mask Transfiner achieves consistently large improvements on different backbones and object detectors. We further observe that the usage of DCN~\cite{zhu2019deformable} with Mask Transfiner can bring a surge in performance. We compare Transfiner with Mask Scoring R-CNN~\cite{huang2019mask} trained with DCN under the same setting and training schedules. Using ResNet-101 and Faster R-CNN~\cite{ren2015faster} detector, the mask AP of Mask Transfiner on COCO \textit{test-dev} is 42.2, while Mask Scoring R-CNN is 39.6 in Table 9 of the paper. For more comprehensive comparisons on two-stage instance segmentation methods, in Table~\ref{tab:dcn_comp}, we also train Mask R-CNN~\cite{he2017mask}, PointRend~\cite{kirillov2020pointrend}, BCNet~\cite{ke2021bcnet}, Cascade Mask R-CNN~\cite{cai2019cascade} and HTC~\cite{chen2019hybrid} with the multi-scale 3$\times$ training schedule with DCN, and submit their predictions to the evaluation server for obtaining their accuracies on the test-dev split. The performance advantages of Mask Transfiner are consistently significant, improving the baseline Mask R-CNN$^\dagger$ for 2.8 mask AP and outperforming PointRend by 0.9 AP.
\vspace{-0.1in}

\begin{table}[!h]
	\caption{Performance comparison between two-stage instance segmentation methods on COCO \textit{test-dev} set using R101-FPN. The dagger $^\dagger$ denotes training with DCN~\cite{zhu2019deformable} and 3$\times$ training schedule in our implementation. HTC and Cascade Mask R-CNN use 3-stage cascade refinement with multiple object detectors and mask heads. The standard transformer with output size 112$\times$112 runs out of memory in our experiments.}
	\vspace{-0.1in}
	\centering
	\resizebox{1.0\linewidth}{!}{
		\begin{tabular}{l | c | c | c | c | c | c}
			\toprule
			Method & Output Size & AP & AP$_{S}$ & AP$_{M}$ & AP$_{L}$ & FPS\\
			\midrule
			Mask R-CNN$^\dagger$~\cite{he2017mask} (Baseline)~\cite{he2017mask} & 28$\times$28 & 39.4 & 18.6 & 42.8 & 54.5 & \textbf{9.6} \\
			Mask Scoring R-CNN$^\dagger$~\cite{huang2019mask} & 28$\times$28 & 39.6 & 18.9 & 42.7 & 55.1 & 9.2 \\
			BCNet$^\dagger$~\cite{ke2021bcnet} & 28$\times$28 & 41.2 & 23.6 & 43.9 & 52.8 & 8.9 \\
			PointRend$^\dagger$~\cite{kirillov2020pointrend} & 224$\times$224 & 41.3 & 20.6 & 44.0 & 55.3 & 7.2 \\
            Cascade Mask R-CNN$^\dagger$~\cite{cai2019cascade} & 28$\times$28 & 41.5 & 22.1 & 42.6 & 54.2 & 4.8 \\
            HTC$^\dagger$~\cite{chen2019hybrid} & 28$\times$28 & 41.7 & 23.3 & 44.2 & 53.8 & 2.1 \\
			\midrule
		    Standard Transformer$^\dagger$ & 56$\times$56 & 41.3 & 23.4 & 43.5 & 53.2 & 1.4 \\
			Mask Transfiner$^\dagger$ (Ours: Quadtree Transformer) & 112$\times$112 & \textbf{42.2} & \textbf{24.1} & \textbf{44.8} & \textbf{55.4} & 6.1 \\
			\bottomrule
		\end{tabular}
	}
	\vspace{-0.25in}
	\label{tab:dcn_comp}
\end{table}

\begin{figure*}[!t]
    \vspace{-0.3in}
	\centering
    \includegraphics[width=0.93\linewidth]{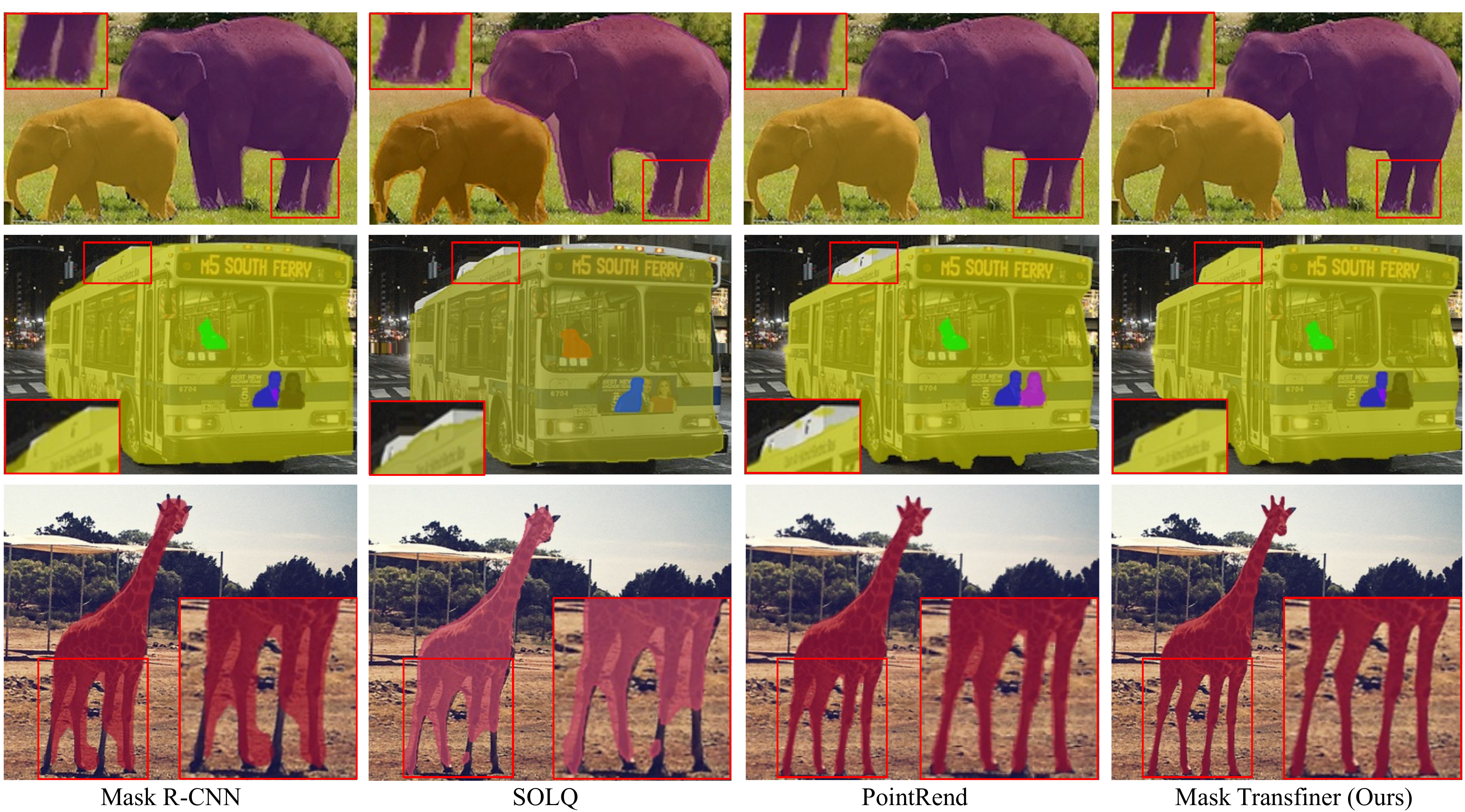}
	\vspace{-0.12in}
	\caption{Instance Segmentation on COCO~\cite{lin2014microsoft} validation set by a) Mask R-CNN~\cite{he2017mask}, b) SOLQ~\cite{dong2021solq}, c) PointRend~\cite{kirillov2020pointrend}, d) Mask Transfiner (Ours) using R50-FPN as backbone, where Mask Transfiner produces significantly more detailed results at high-frequency image regions by replacing Mask R-CNN's default mask head. Zoom in for better view.}
    \label{fig:coco_comp}
    \vspace{-0.1in}
\end{figure*}

\begin{figure*}[!h]
	\centering
    \includegraphics[width=0.97\linewidth]{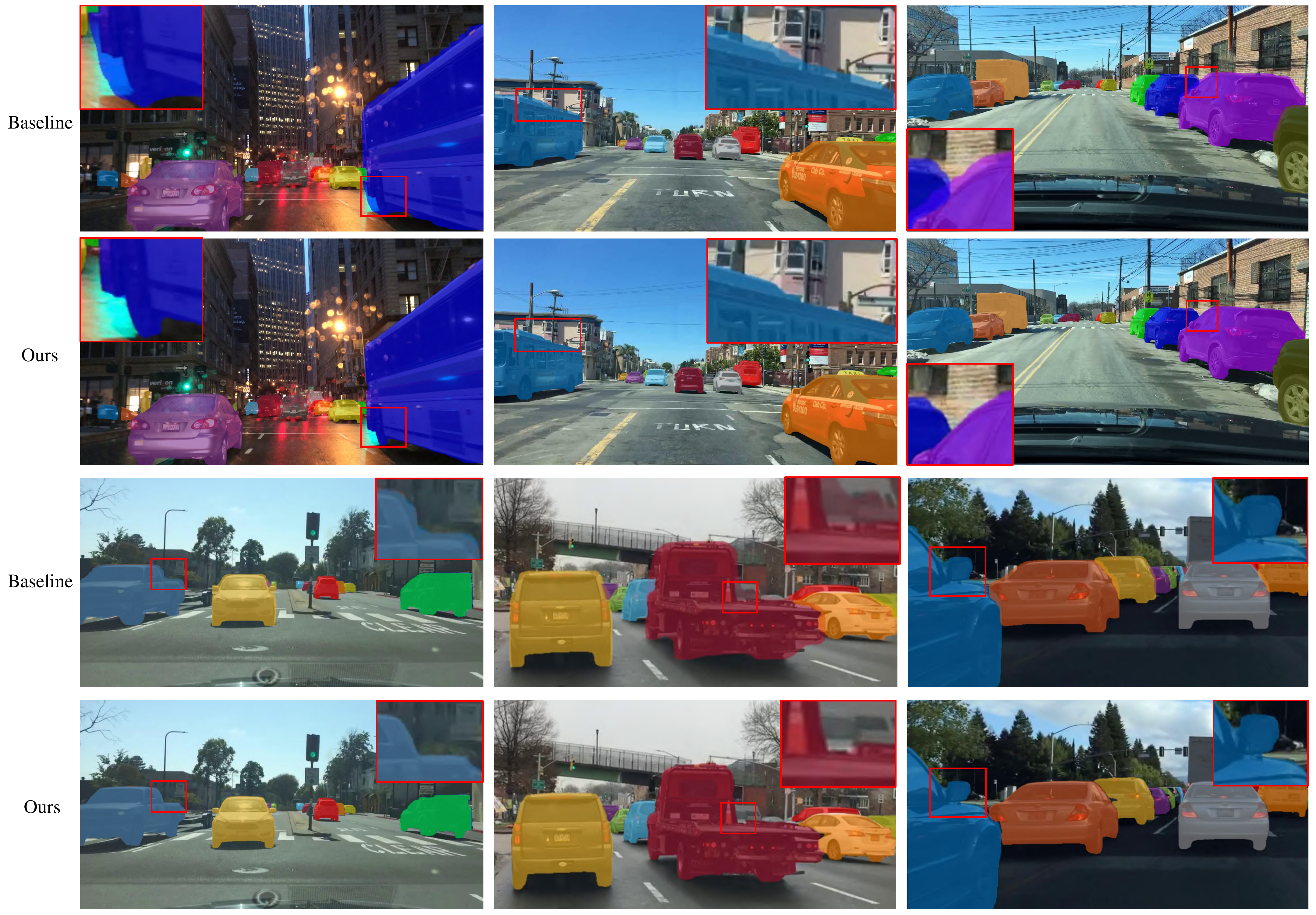}
	\vspace{-0.1in}
	\caption{Qualitative comparisons with baseline method Mask R-CNN~\cite{he2017mask} and our Mask Transfiner on BDD100K~\cite{bdd100k} \textit{val} set. Mask Transfiner produces more correct and natural segmentation results by revealing details for high-frequency regions.}
    \label{fig:bdd_comp}
    \vspace{-0.2in}
\end{figure*}

\begin{figure*}[!h]
	\centering
    \includegraphics[width=1.0\linewidth]{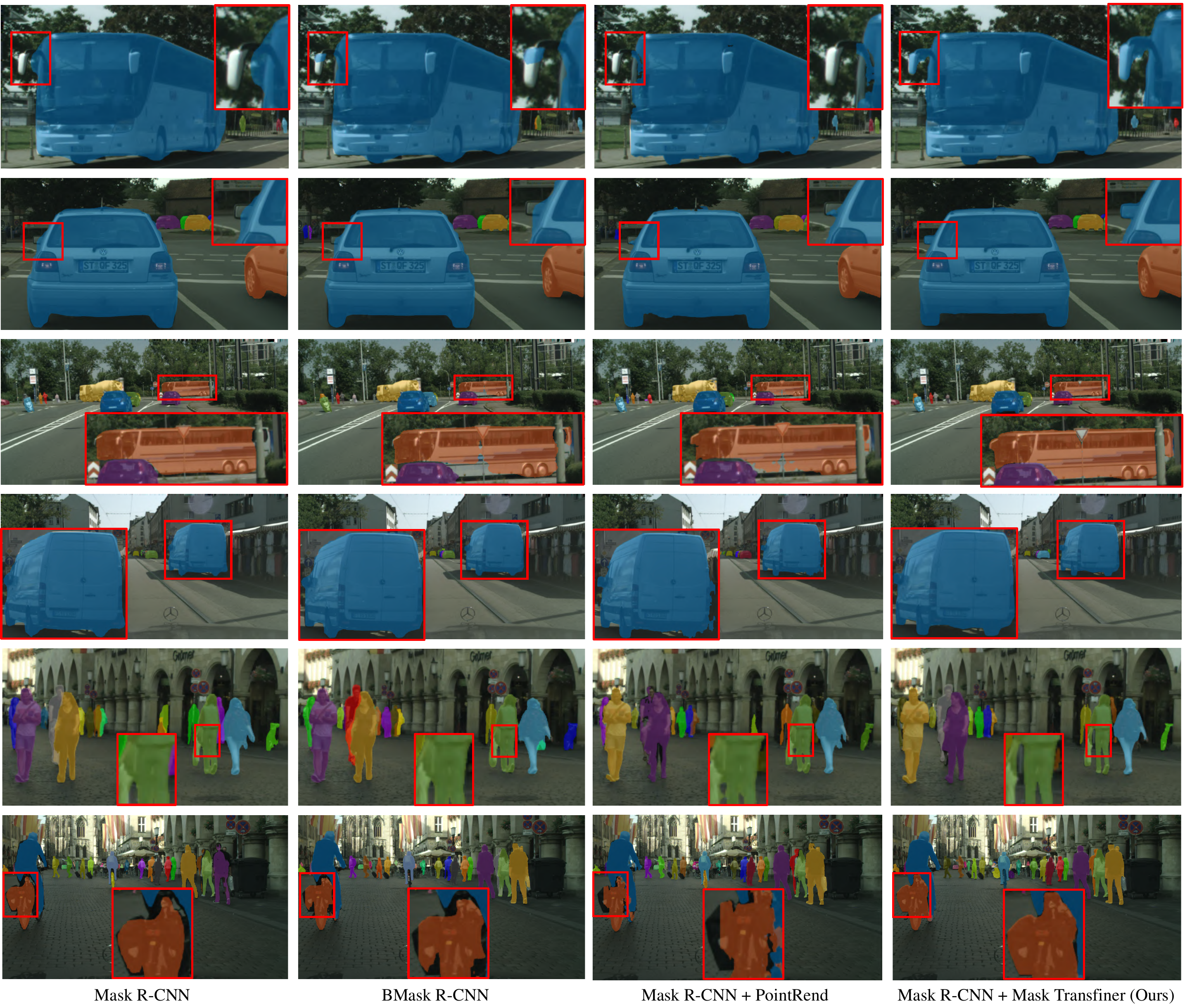}
	\vspace{-0.28in}
	\caption{Qualitative comparisons with instance segmentation methods Mask R-CNN~\cite{he2017mask}, BMask R-CNN~\cite{ChengWHL20}, PointRend~\cite{kirillov2020pointrend} and our Mask Transfiner on Cityscapes~\cite{cordts2016cityscapes} \textit{val} set. Mask Transfiner produces more precise and natural segmentation results, where even the small triangle-shaped traffic sign occluding the bus (3rd row) and the gap between the hand and leg (5th row) could be correctly separated.}
    \label{fig:city_comp}
    \vspace{-0.2in}
\end{figure*}

\paragraph{Inference Speed} We adopt frames per second (FPS) to evaluate the inference speed of the models. In Table~\ref{tab:dcn_comp}, we benchmark all the compared two-stage methods using a Titan RTX GPU. The reported FPS is the average obtained in five runs, where each run measures the FPS of a model through 200 iterations. Compared to the Cascade Mask R-CNN and HTC with three-stage cascade refinement and multiple object detectors/mask heads (output size 28$\times$28), our Transfiner using 3-level quadtree is much faster and more accurate with higher-resolution predictions (112$\times$112). Comparing to the baseline Mask R-CNN, although there is a drop on inference speed for about 35\% due to multi-head attention modeling between hierarchical incoherent regions, the significant performance boost of 2.8 mask AP and 4 times larger output height/width are good compensation trade-offs. Note that standard transformer (3 layers and 4 attention heads in each layer) operating on uniform grids with output size 56$\times$56 only runs at 1.4 FPS, which is much slower than our method. 

\begin{figure*}[!t]
    \vspace{-0.30in}
	\centering
	\includegraphics[width=1.0\linewidth]{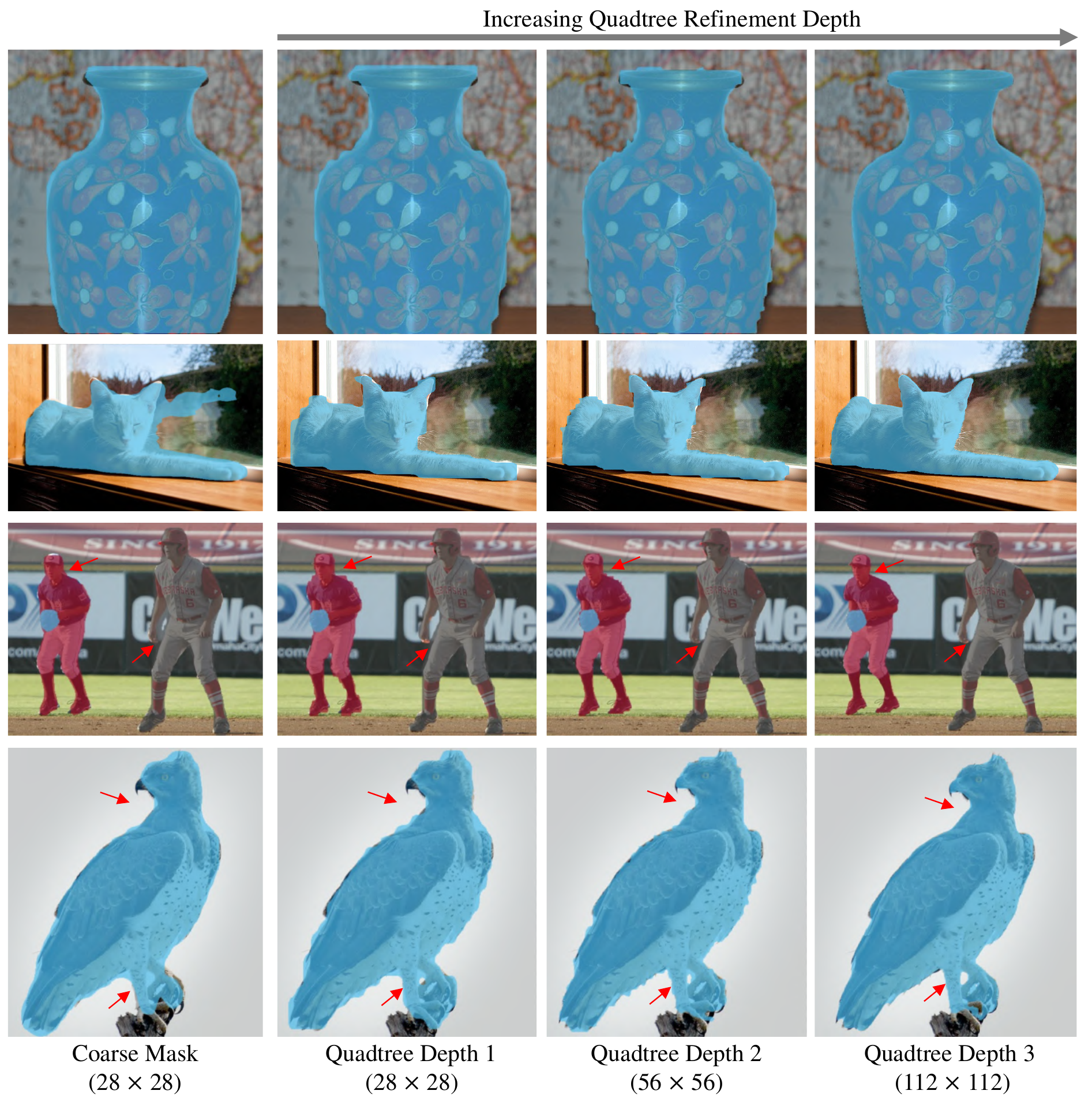}
	\vspace{-0.28in}
	\caption{Qualitative results comparison between the coarse mask predictions by our baseline~\cite{he2016deep} and the refinement results of Mask Transfiner on COCO with various depths of the quadtree built on detected incoherent regions.}
	\label{fig:refine_process}
	\vspace{-0.15in}
\end{figure*}

\subsection{More Qualitative Comparisons}
\label{sec:qual_comp}

We provide more qualitative results comparisons on three evaluation benchmarks COCO  (Figure~\ref{fig:coco_comp}), B100K (Figure~\ref{fig:bdd_comp}) and Cityscapes (Figure~\ref{fig:city_comp}), where our Mask Transfiner consistently produces masks with substantially higher precision and quality than previous methods~\cite{he2017mask, dong2021solq,ChengWHL20, kirillov2020pointrend}. Take the third case in Figure~\ref{fig:coco_comp} as an example, SOLQ and the baseline Mask R-CNN 
only provides very coarse mask predictions in the high-frequency regions, such as the giraffe's head and feet regions, due to their low-resolution output sizes 28$\times$28. Although PointRend employs 
large output size 224$\times$224, it still fails to delineate the thin gap between the left legs of giraffe. Note that the mask output size of Mask Transfiner only is 112$\times$112. These segmentation errors on ambiguous regions reveal the limitation of segmenting each pixel separately only by a share MLP~\cite{kirillov2020pointrend} without global reasoning and hierarchical pixel-wise relations modeling.


\subsection{Visual Analysis}
\label{sec:vis_analysis}

\paragraph{Visualization Multi-level Refinement}
In Figure~\ref{fig:refine_process}, we analyze how the mask predictions evolve with increasing quadtree depths. The predicted masks become substantially finer in detail around object boundaries, which reveals that the quadtree nodes with more levels at larger output sizes for an object preserves more low-level details for fine-grained segmentation.
\vspace{-0.15in}

\begin{figure*}[!t]
    \vspace{-0.32in}
	\centering
	\includegraphics[width=1.0\linewidth]{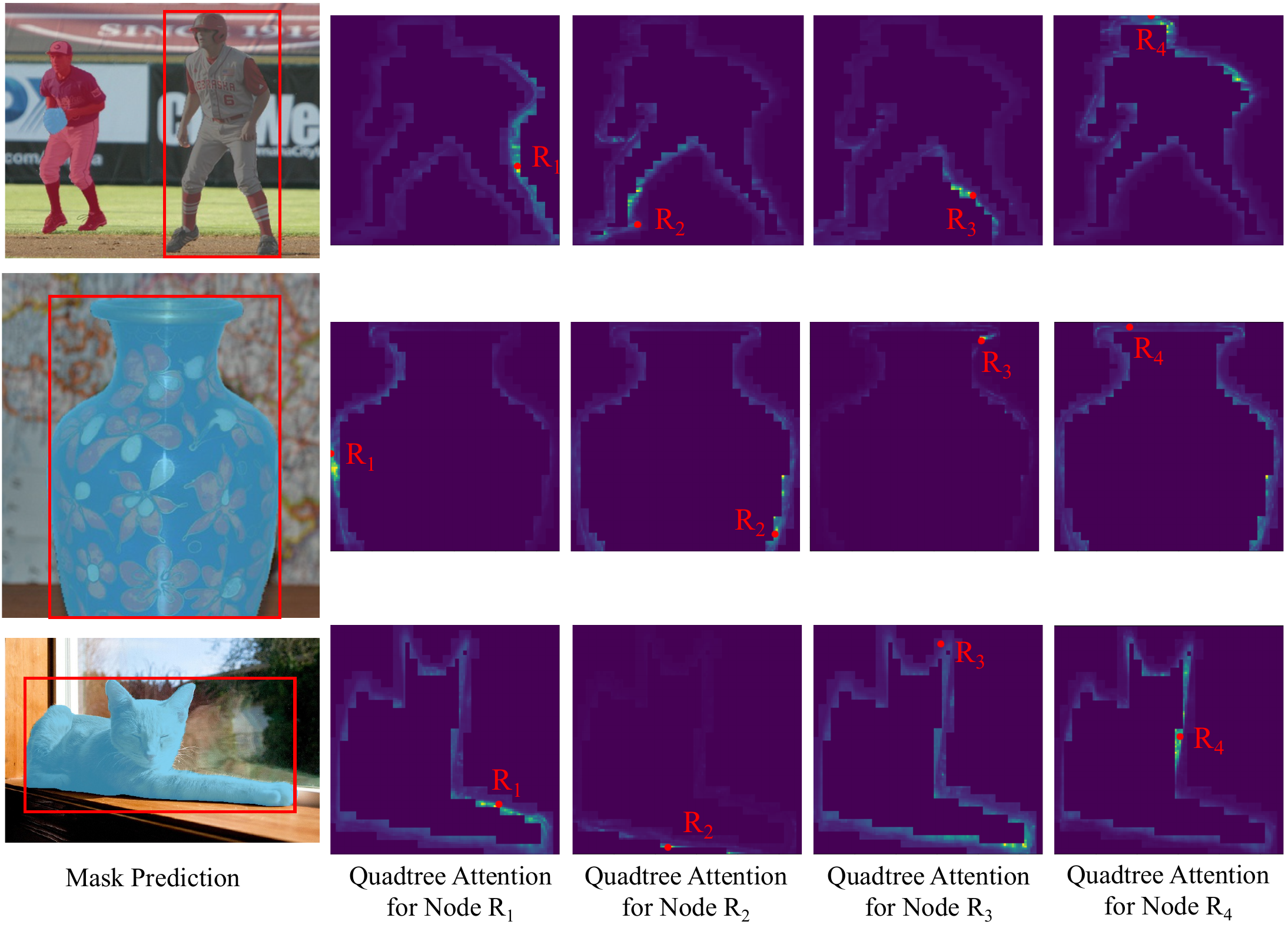}
	\vspace{-0.27in}
	\caption{Visualization on the quadtree attention weights distribution in the sparse incoherent regions for four sampled red nodes.}
	\label{fig:att_inc}
	\vspace{-0.1in}
\end{figure*}

\begin{figure*}[!h]
	\centering
	\includegraphics[width=1.0\linewidth]{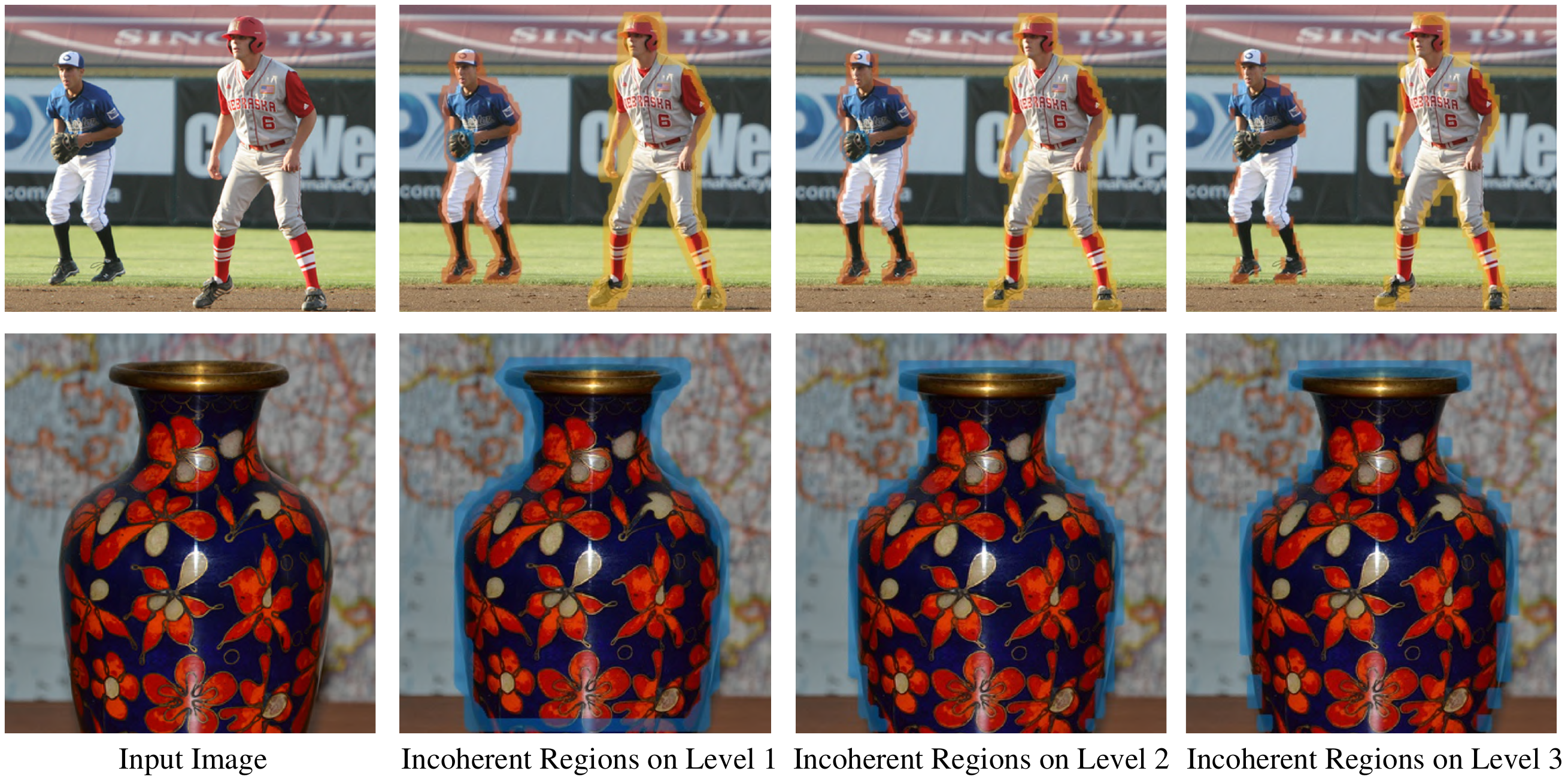}
	\vspace{-0.27in}
	\caption{Visualization of detected incoherent regions on different levels of the constructed quadtree based on the RoI pyramid, where the incoherent nodes regions in deeper quadtree levels with larger resolution size are distributed more sparsely.}
	\label{fig:inc_region}
	\vspace{-0.15in}
\end{figure*}

\paragraph{Failure Cases} We also analyze the failure cases and find one typical failure mode shown in the last row of Figure~\ref{fig:refine_process}, where a small portion of the bird's paw is wrongly predicted as background wood due to their highly similar appearance and texture.

\vspace{-0.2in}
\paragraph{Visualization on Quadtree Attention and Incoherent Regions}
In Figure~\ref{fig:att_inc} and Figure~\ref{fig:inc_region},
we further provide more quadtree attention visualization examples and their detected incoherent regions on RoI pyramid, where the outline of objects can be observed and the sparsity of quadtree attention is clearly shown. The quadtree nodes with higher appearance or positional similarity has larger attention weights attending between them.


{\small
\bibliographystyle{ieee_fullname}
\bibliography{egbib}

\begin{thebibliography}{10}\itemsep=-1pt

\bibitem{bolya2019yolact}
Daniel Bolya, Chong Zhou, Fanyi Xiao, and Yong~Jae Lee.
\newblock Yolact: real-time instance segmentation.
\newblock In {\em ICCV}, 2019.

\bibitem{cai2018cascade}
Zhaowei Cai and Nuno Vasconcelos.
\newblock Cascade r-cnn: Delving into high quality object detection.
\newblock In {\em CVPR}, 2018.

\bibitem{cai2019cascade}
Zhaowei Cai and Nuno Vasconcelos.
\newblock Cascade r-cnn: High quality object detection and instance
  segmentation.
\newblock 2019.

\bibitem{carion2020end}
Nicolas Carion, Francisco Massa, Gabriel Synnaeve, Nicolas Usunier, Alexander
  Kirillov, and Sergey Zagoruyko.
\newblock End-to-end object detection with transformers.
\newblock In {\em ECCV}, 2020.

\bibitem{chen2020blendmask}
Hao Chen, Kunyang Sun, Zhi Tian, Chunhua Shen, Yongming Huang, and Youliang
  Yan.
\newblock {BlendMask}: Top-down meets bottom-up for instance segmentation.
\newblock In {\em CVPR}, 2020.

\bibitem{chen2019hybrid}
Kai Chen, Jiangmiao Pang, Jiaqi Wang, Yu Xiong, Xiaoxiao Li, Shuyang Sun,
  Wansen Feng, Ziwei Liu, Jianping Shi, Wanli Ouyang, et~al.
\newblock Hybrid task cascade for instance segmentation.
\newblock In {\em CVPR}, 2019.

\bibitem{chen2018masklab}
Liang-Chieh Chen, Alexander Hermans, George Papandreou, Florian Schroff, Peng
  Wang, and Hartwig Adam.
\newblock Masklab: Instance segmentation by refining object detection with
  semantic and direction features.
\newblock In {\em CVPR}, 2018.

\bibitem{chen2019tensormask}
Xinlei Chen, Ross Girshick, Kaiming He, and Piotr Doll{\'a}r.
\newblock Tensormask: A foundation for dense object segmentation.
\newblock In {\em ICCV}, 2019.

\bibitem{cheng2020panoptic}
Bowen Cheng, Maxwell~D Collins, Yukun Zhu, Ting Liu, Thomas~S Huang, Hartwig
  Adam, and Liang-Chieh Chen.
\newblock Panoptic-deeplab: A simple, strong, and fast baseline for bottom-up
  panoptic segmentation.
\newblock In {\em CVPR}, 2020.

\bibitem{cheng2021boundary}
Bowen Cheng, Ross Girshick, Piotr Doll{\'a}r, Alexander~C Berg, and Alexander
  Kirillov.
\newblock Boundary iou: Improving object-centric image segmentation evaluation.
\newblock In {\em CVPR}, 2021.

\bibitem{cheng2020cascadepsp}
Ho~Kei Cheng, Jihoon Chung, Yu-Wing Tai, and Chi-Keung Tang.
\newblock Cascadepsp: toward class-agnostic and very high-resolution
  segmentation via global and local refinement.
\newblock In {\em CVPR}, 2020.

\bibitem{ChengWHL20}
Tianheng Cheng, Xinggang Wang, Lichao Huang, and Wenyu Liu.
\newblock Boundary-preserving mask r-cnn.
\newblock In {\em ECCV}, 2020.

\bibitem{cordts2016cityscapes}
Marius Cordts, Mohamed Omran, Sebastian Ramos, Timo Rehfeld, Markus Enzweiler,
  Rodrigo Benenson, Uwe Franke, Stefan Roth, and Bernt Schiele.
\newblock The cityscapes dataset for semantic urban scene understanding.
\newblock In {\em CVPR}, 2016.

\bibitem{dong2021solq}
Bin Dong, Fangao Zeng, Tiancai Wang, Xiangyu Zhang, and Yichen Wei.
\newblock Solq: Segmenting objects by learning queries.
\newblock In {\em NeurIPS}, 2021.

\bibitem{QueryInst}
Yuxin Fang, Shusheng Yang, Xinggang Wang, Yu Li, Chen Fang, Ying Shan, Bin
  Feng, and Wenyu Liu.
\newblock Instances as queries.
\newblock In {\em ICCV}, 2021.

\bibitem{finkel1974quad}
Raphael~A Finkel and Jon~Louis Bentley.
\newblock Quad trees a data structure for retrieval on composite keys.
\newblock {\em Acta informatica}, 4(1):1--9, 1974.

\bibitem{guo2021sotr}
Ruohao Guo, Dantong Niu, Liao Qu, and Zhenbo Li.
\newblock Sotr: Segmenting objects with transformers.
\newblock In {\em ICCV}, 2021.

\bibitem{gupta2019lvis}
Agrim Gupta, Piotr Dollar, and Ross Girshick.
\newblock Lvis: A dataset for large vocabulary instance segmentation.
\newblock In {\em CVPR}, 2019.

\bibitem{he2017mask}
Kaiming He, Georgia Gkioxari, Piotr Doll{\'a}r, and Ross Girshick.
\newblock Mask r-cnn.
\newblock In {\em ICCV}, 2017.

\bibitem{he2016deep}
Kaiming He, Xiangyu Zhang, Shaoqing Ren, and Jian Sun.
\newblock Deep residual learning for image recognition.
\newblock In {\em CVPR}, 2016.

\bibitem{hu2021ISTR}
Jie Hu, Liujuan Cao, Yao Lu, ShengChuan Zhang, Ke Li, Feiyue Huang, Ling Shao,
  and Rongrong Ji.
\newblock Istr: End-to-end instance segmentation via transformers.
\newblock {\em arXiv preprint arXiv:2105.00637}, 2021.

\bibitem{huang2019mask}
Zhaojin Huang, Lichao Huang, Yongchao Gong, Chang Huang, and Xinggang Wang.
\newblock Mask scoring r-cnn.
\newblock In {\em CVPR}, 2019.

\bibitem{pcan}
Lei Ke, Xia Li, Martin Danelljan, Yu-Wing Tai, Chi-Keung Tang, and Fisher Yu.
\newblock Prototypical cross-attention networks for multiple object tracking
  and segmentation.
\newblock In {\em NeurIPS}, 2021.

\bibitem{ke2021bcnet}
Lei Ke, Yu-Wing Tai, and Chi-Keung Tang.
\newblock Deep occlusion-aware instance segmentation with overlapping bilayers.
\newblock In {\em CVPR}, 2021.

\bibitem{kirillov2020pointrend}
Alexander Kirillov, Yuxin Wu, Kaiming He, and Ross Girshick.
\newblock Pointrend: Image segmentation as rendering.
\newblock In {\em CVPR}, 2020.

\bibitem{kuo2019shapemask}
Weicheng Kuo, Anelia Angelova, Jitendra Malik, and Tsung-Yi Lin.
\newblock Shapemask: Learning to segment novel objects by refining shape
  priors.
\newblock In {\em ICCV}, 2019.

\bibitem{lee2019centermask}
Youngwan Lee and Jongyoul Park.
\newblock Centermask: Real-time anchor-free instance segmentation.
\newblock In {\em CVPR}, 2020.

\bibitem{li2017fully}
Yi Li, Haozhi Qi, Jifeng Dai, Xiangyang Ji, and Yichen Wei.
\newblock Fully convolutional instance-aware semantic segmentation.
\newblock In {\em CVPR}, 2017.

\bibitem{liang2020polytransform}
Justin Liang, Namdar Homayounfar, Wei-Chiu Ma, Yuwen Xiong, Rui Hu, and Raquel
  Urtasun.
\newblock Polytransform: Deep polygon transformer for instance segmentation.
\newblock In {\em CVPR}, 2020.

\bibitem{lin2017feature}
Tsung-Yi Lin, Piotr Doll{\'a}r, Ross Girshick, Kaiming He, Bharath Hariharan,
  and Serge Belongie.
\newblock Feature pyramid networks for object detection.
\newblock In {\em CVPR}, 2017.

\bibitem{lin2014microsoft}
Tsung-Yi Lin, Michael Maire, Serge Belongie, James Hays, Pietro Perona, Deva
  Ramanan, Piotr Doll{\'a}r, and C~Lawrence Zitnick.
\newblock Microsoft coco: Common objects in context.
\newblock In {\em ECCV}, 2014.

\bibitem{liu2018path}
Shu Liu, Lu Qi, Haifang Qin, Jianping Shi, and Jiaya Jia.
\newblock Path aggregation network for instance segmentation.
\newblock In {\em CVPR}, 2018.

\bibitem{RSLoss}
Kemal Oksuz, Baris~Can Cam, Emre Akbas, and Sinan Kalkan.
\newblock Rank sort loss for object detection and instance segmentation.
\newblock In {\em ICCV}, 2021.

\bibitem{ren2015faster}
Shaoqing Ren, Kaiming He, Ross Girshick, and Jian Sun.
\newblock Faster r-cnn: Towards real-time object detection with region proposal
  networks.
\newblock In {\em NeurIPS}, 2015.

\bibitem{takikawa2019gated}
Towaki Takikawa, David Acuna, Varun Jampani, and Sanja Fidler.
\newblock Gated-scnn: Gated shape cnns for semantic segmentation.
\newblock In {\em ICCV}, 2019.

\bibitem{tang2021look}
Chufeng Tang, Hang Chen, Xiao Li, Jianmin Li, Zhaoxiang Zhang, and Xiaolin Hu.
\newblock Look closer to segment better: Boundary patch refinement for instance
  segmentation.
\newblock 2021.

\bibitem{vaswani2017attention}
Ashish Vaswani, Noam Shazeer, Niki Parmar, Jakob Uszkoreit, Llion Jones,
  Aidan~N Gomez, {\L}ukasz Kaiser, and Illia Polosukhin.
\newblock Attention is all you need.
\newblock In {\em NeurIPS}, 2017.

\bibitem{wang2020deep}
Jingdong Wang, Ke Sun, Tianheng Cheng, Borui Jiang, Chaorui Deng, Yang Zhao,
  Dong Liu, Yadong Mu, Mingkui Tan, Xinggang Wang, et~al.
\newblock Deep high-resolution representation learning for visual recognition.
\newblock {\em TPAMI}, 2020.

\bibitem{wang2018non}
Xiaolong Wang, Ross Girshick, Abhinav Gupta, and Kaiming He.
\newblock Non-local neural networks.
\newblock In {\em CVPR}, 2018.

\bibitem{wang2019solo}
Xinlong Wang, Tao Kong, Chunhua Shen, Yuning Jiang, and Lei Li.
\newblock Solo: Segmenting objects by locations.
\newblock {\em arXiv preprint arXiv:1912.04488}, 2019.

\bibitem{wang2020solov2}
Xinlong Wang, Rufeng Zhang, Tao Kong, Lei Li, and Chunhua Shen.
\newblock Solov2: Dynamic and fast instance segmentation.
\newblock In {\em NeurIPS}, 2020.

\bibitem{wang2020end}
Yuqing Wang, Zhaoliang Xu, Xinlong Wang, Chunhua Shen, Baoshan Cheng, Hao Shen,
  and Huaxia Xia.
\newblock End-to-end video instance segmentation with transformers.
\newblock In {\em CVPR}, 2021.

\bibitem{wu2019detectron2}
Yuxin Wu, Alexander Kirillov, Francisco Massa, Wan-Yen Lo, and Ross Girshick.
\newblock Detectron2.
\newblock \url{https://github.com/facebookresearch/detectron2}, 2019.

\bibitem{xie2019polarmask}
Enze Xie, Peize Sun, Xiaoge Song, Wenhai Wang, Xuebo Liu, Ding Liang, Chunhua
  Shen, and Ping Luo.
\newblock Polarmask: Single shot instance segmentation with polar
  representation.
\newblock In {\em CVPR}, 2020.

\bibitem{bdd100k}
Fisher Yu, Haofeng Chen, Xin Wang, Wenqi Xian, Yingying Chen, Fangchen Liu,
  Vashisht Madhavan, and Trevor Darrell.
\newblock Bdd100k: A diverse driving dataset for heterogeneous multitask
  learning.
\newblock In {\em CVPR}, 2020.

\bibitem{yuan2020segfix}
Yuhui Yuan, Jingyi Xie, Xilin Chen, and Jingdong Wang.
\newblock Segfix: Model-agnostic boundary refinement for segmentation.
\newblock In {\em ECCV}, 2020.

\bibitem{refinemask}
Gang Zhang, Xin Lu, Jingru Tan, Jianmin Li, Zhaoxiang Zhang, Quanquan Li, and
  Xiaolin Hu.
\newblock Refinemask: Towards high-quality instance segmentation with
  fine-grained features.
\newblock In {\em CVPR}, 2021.

\bibitem{zhang2021k}
Wenwei Zhang, Jiangmiao Pang, Kai Chen, and Chen~Change Loy.
\newblock K-net: Towards unified image segmentation.
\newblock In {\em NeurIPS}, 2021.

\bibitem{zhu2019deformable}
Xizhou Zhu, Han Hu, Stephen Lin, and Jifeng Dai.
\newblock Deformable convnets v2: More deformable, better results.
\newblock In {\em CVPR}, 2019.

\end{thebibliography}
}

\end{document}